\documentclass{article}

\PassOptionsToPackage{numbers, compress}{natbib}

\usepackage[main, final]{neurips_2026}

\usepackage[utf8]{inputenc} 
\usepackage[T1]{fontenc}    
\usepackage{hyperref}       
\usepackage{url}            
\usepackage{booktabs}       
\usepackage{amsfonts}       
\usepackage{nicefrac}       
\usepackage{microtype}      
\usepackage{xcolor}         
\usepackage{graphicx}
\usepackage{booktabs}
\usepackage{multirow} 
\usepackage{adjustbox}
\usepackage{amsmath}
\usepackage{subcaption}   
\usepackage{amssymb}   
\usepackage{pifont}    
\providecommand{\gain}[1]{{\scriptsize(#1)}}
\providecommand{\bestgain}[1]{{\scriptsize(#1)}$^{\dagger}$}
\providecommand{\secondgain}[1]{{\scriptsize(#1)}$^{\ddagger}$}

\title{LucidNFT: LR-Anchored Multi-Reward Preference Optimization for Flow-Based Real-World Super-Resolution}

\usepackage{xcolor}
\usepackage{listings}
\usepackage[most]{tcolorbox}
\usepackage{enumitem}
\definecolor{best}{rgb}{1.0, 0.85, 0.6}      
\definecolor{second}{rgb}{0.7, 0.9, 1.0}     
\definecolor{sh_blue}{rgb}{0,0.60,0.93}
\definecolor{sh_gray2}{rgb}{1,0.89,0.75}
\definecolor{lyellow}{rgb}{1,0.63,0.098}
\definecolor{lred}{rgb}{0.906,0.42,0.32}
\definecolor{color3}{rgb}{0.95,0.95,0.95}
\definecolor{mygray}{gray}{.9}
\definecolor{genhaze}{rgb}{0.60, 0.57, 0.79}
\definecolor{bluegreen}{rgb}{0.44, 0.64, 0.77}
\definecolor{gray_venue}{rgb}{0.53,0.52,0.52}
\definecolor{color5}{rgb}{1,0.96,0.88}

\lstdefinelanguage{json}{
  basicstyle=\ttfamily\small,
  breaklines=true,
  keepspaces=true,
  showstringspaces=false,
  morecomment=[l]{//},
  morestring=[b]",
  literate=
    *{0}{{{\color{black}{0}}}}1
     {1}{{{\color{black}{1}}}}1
     {:}{{{\color{black}{:}}}}1
     {,}{{{\color{black}{,}}}}1
     {\{}{{{\color{black}{\{}}}}1
     {\}}{{{\color{black}{\}}}}}1
     {[}{{{\color{black}{[}}}}1
     {]}{{{\color{black}{]}}}}1
}

\tcbuselibrary{listingsutf8,skins}
\tcbset{
    examplebox/.style={
        enhanced,
        colback=sh_blue!5,          
        colframe=sh_blue,           
        coltitle=sh_blue,           
        breakable,   
        fonttitle=\bfseries,
        boxrule=0.6mm,
        sharp corners,
        attach boxed title to top left={
            yshift=-2mm,
            xshift=5mm
        },
        boxed title style={
            colframe=sh_blue,       
            colback=white,          
            sharp corners,
            boxrule=0.6mm
        },
        before upper={\small\rmfamily\color{black}}, 
        fontupper=\small,
        boxsep=6pt,
        left=10pt,
        right=10pt,
        top=8pt,
        bottom=8pt
    }
}

\newtcolorbox[auto counter, number within=section]{promptbox}[2][]{%
    examplebox,
    title=Prompt~\thetcbcounter~(#2),
    #1
}
\newtcolorbox[auto counter, number within=section]{algobox}[2][]{%
    examplebox,
    title=Algorithm~\thetcbcounter~(#2),
    #1
}

\author{%
  Song Fei$^{1\dagger}$ \quad Tian Ye$^{1\dagger}$ \quad Sixiang Chen$^{1}$ \quad Zhaohu Xing$^{1}$ \quad Jianyu Lai$^{1}$ \quad Lei Zhu$^{1,2}$\thanks{Lei Zhu is the corresponding author of this work.} \\
  \vspace{2pt}
  $^{1}$The Hong Kong University of Science and Technology (Guangzhou) \\
  $^{2}$The Hong Kong University of Science and Technology \\
  \vspace{2pt}
  {\small $^{\dagger}$Equal contribution } \\
  \vspace{2pt}
  {\small \textbf{Project page:} \url{https://w2genai-lab.github.io/LucidNFT}} \\ {\small \textbf{Code:} \url{https://github.com/W2GenAI-Lab/LucidNFT}}
}

\begin{document}

\maketitle

\begin{abstract}
Generative real-world image super-resolution (Real-ISR) can synthesize visually convincing details from severely degraded low-resolution (LR) inputs, yet its stochastic sampling makes a critical failure mode hard to avoid: outputs may look sharp but be \emph{unfaithful} to the LR evidence, exhibiting semantic or structural hallucinations. Once a generative prior has been adapted to Real-ISR, its stochastic sampling turns each LR input into a \emph{rollout group} of candidates with different perceptual--faithfulness trade-offs, which allows reinforcement learning (RL) post-training to optimize objectives that paired HR supervision cannot specify. However, effective alignment in Real-ISR is hindered by three coupled challenges: (i) the lack of an LR-referenced faithfulness signal that is robust to degradation yet sensitive to localized hallucinations, (ii) a rollout-group optimization bottleneck where scalarizing heterogeneous rewards before normalization compresses objective-wise contrasts and weakens the reward-weighted updates of Diffusion Negative-aware FineTuning (DiffusionNFT), and (iii) limited coverage of real degradations, which restricts rollout diversity and preference signal quality. We propose \textbf{LucidNFT}, a multi-reward RL framework for \emph{flow-matching} Real-ISR. LucidNFT introduces \textbf{LucidConsistency}, a degradation-invariant and hallucination-sensitive LR-referenced evaluator trained with content-consistent degradation pools and original--inpainted hard negatives; a \textbf{decoupled reward normalization} strategy that preserves objective-wise contrasts within each LR-conditioned rollout group before fusion; and \textbf{LucidLR}, a large-scale collection of real-world degraded images for robust RL fine-tuning. Extensive experiments show that LucidNFT improves perceptual quality on strong flow-based Real-ISR baselines while generally maintaining LR-referenced consistency across diverse real-world scenarios, and a reward-independent human study judges its outputs more LR-faithful in 72.5\% of comparisons.
\end{abstract}

\vspace{-6mm}
\section{Introduction}
\label{sec:intro}
\vspace{-2mm}

Real-world image super-resolution (Real-ISR) aims to recover a high-resolution (HR) image from a degraded low-resolution (LR) observation under \emph{unknown} and \emph{heterogeneous} degradations. Empowered by large-scale generative priors, recent generative Real-ISR approaches---including diffusion-based methods~\cite{StableSR,Resshift,SinSR,SeeSR,DiffBIR,DreamClear,SUPIR} and scalable flow-matching formulations~\cite{LucidFlux,DiT4SR}---have substantially improved perceptual restoration quality by synthesizing rich and plausible high-frequency details beyond regression-based frameworks~\cite{zhang2018image,dong2014learning,chen2023activating}. 
However, this generative freedom also exposes a central reliability bottleneck: restored details may look realistic yet be \emph{unfaithful} to the LR evidence (i.e., semantic or structural hallucination~\cite{cohen2024looks}). In the absence of HR ground truth, there is no reliable way to \emph{measure}, and therefore to \emph{optimize}, such LR-anchored faithfulness, which is a key obstacle to deploying and aligning generative Real-ISR models in real-world settings.

Stochastic sampling is not only the source of hallucination but also a source of supervision. Across image generation~\cite{Lens}, video generation~\cite{Seedance}, and language modeling~\cite{InstructGPT}, a standard recipe first establishes basic capability through supervised training and then applies reinforcement learning (RL)~\cite{RL} to optimize objectives that direct supervision cannot easily specify. Real-ISR fits this recipe: backbones such as LucidFlux~\cite{LucidFlux} and DiT4SR~\cite{DiT4SR} have already adapted generative priors to restoration, and for a fixed LR input their stochastic sampling yields candidates with different perceptual--faithfulness trade-offs, some adding plausible but LR-unsupported details and others preserving the LR evidence at the cost of visual appeal. Large real LR collections also lack paired HR references, so paired supervision cannot exploit them, whereas rewards on generated candidates can. Preference-based RL post-training is therefore a principled, though not the only, route to promoting candidates that are both perceptually strong and LR-faithful.

Making preference optimization effective for Real-ISR, however, is challenging. Preference learning compares rollouts \emph{conditioned on the same LR input}, and these candidates can excel in different objectives. The common practice of scalarizing multiple rewards into one score before normalization~\cite{DiffusionNFT,Flow-GRPO,DanceGRPO} allows the reward with the larger within-group variance to dominate the advantage, so that candidates with distinct perceptual--faithfulness trade-offs receive similar advantages. We build on Diffusion Negative-aware FineTuning (DiffusionNFT)~\cite{DiffusionNFT}, a forward-process RL objective for flow-matching models that is compatible with black-box solvers, is reported to be $3$--$25\times$ more efficient than FlowGRPO~\cite{Flow-GRPO}, and has been adopted for image editing, few-step generation, and unified RL toolkits~\cite{UniWorld-V2,FireRed-Image-Edit,MeanFlowNFT,Flow-Factory}. Like the group-based methods above, DiffusionNFT consumes a single scalar reward, and therefore inherits this \emph{advantage collapse} whenever several rewards are fused, in which case the faithfulness reward can be outweighed by the perceptual one.

Two further conditions are needed for RL alignment to be meaningful. First, no-reference perceptual metrics~\cite{MUSIQ,CLIP-IQA,Q-Align,NIMA,NIQE,MANIQA} do not evaluate LR-anchored faithfulness, and optimizing them alone can reward over-sharpening and hallucinated textures; Real-ISR needs an \emph{LR-referenced} faithfulness signal that is robust to unknown degradations. Second, RL relies on informative rollouts, but paired Real-ISR datasets~\cite{RealSR,DRealSR} are small and synthetic degradation pipelines~\cite{Real-ESRGAN,DIV2K_and_Flickr2K} do not capture entangled real-world degradations.

To address these challenges, we propose LucidNFT, a multi-reward RL framework for aligning flow-matching Real-ISR models. LucidNFT consists of three components:
\begin{itemize}
    \item \textbf{LucidConsistency}, a degradation-invariant and hallucination-sensitive LR-referenced evaluator, trained with multi-degradation views as positives and original--inpainted pairs as hard negatives.
    \item \textbf{Decoupled reward normalization}, which normalizes each reward within its LR-conditioned rollout group before fusion, so that neither reward dominates the other.
    \item \textbf{LucidLR}, a collection of 20K real-world degraded images that provides diverse LR inputs for informative rollouts, released with per-image license metadata.
\end{itemize}

LucidNFT does not introduce a new policy objective: it keeps the DiffusionNFT update unchanged and redesigns the step that maps a multi-dimensional reward vector to the scalar reward weight that DiffusionNFT consumes. Experiments on two flow-based Real-ISR backbones show that LucidNFT improves perceptual quality while generally maintaining LR-referenced consistency, and a human study that does not rely on LucidConsistency prefers LucidNFT outputs as more LR-faithful in 72.5\% of judgments.

\vspace{-3mm}
\section{Related Work}
\label{related}
\vspace{-3mm}

\subsection{Generative Real-World Image Super-Resolution}
\vspace{-2mm}
Recent Real-ISR methods increasingly rely on large-scale generative priors. Diffusion-based approaches~\cite{Resshift,StableSR,SeeSR,DiffBIR,DreamClear,SUPIR} leverage pretrained text-to-image models~\cite{SD,SDXL,SD3,PixArt-alpha,FLUX} to synthesize perceptually rich details beyond regression-based SR, while flow-based formulations~\cite{LucidFlux,DiT4SR} improve scalability and sampling efficiency through velocity-field learning.

\vspace{-3mm}
\subsection{Evaluation Methods for Real-World Image Super-Resolution}
\vspace{-2mm}
In the absence of HR references, full-reference metrics such as PSNR and SSIM~\cite{PSNR} are inapplicable to real-world settings, and Real-ISR is commonly evaluated with no-reference image quality assessment~\cite{MUSIQ,NIMA,MANIQA,NIQE}, including CLIP-based and vision--language evaluators~\cite{CLIP-IQA,Q-Align}. These metrics measure perceptual quality but not whether details are supported by the LR input (Sec.~\ref{sec:lucidconsistency}).

\vspace{-3mm}
\subsection{Reinforcement Learning for Vision Generation Models}
\vspace{-2mm}
Reinforcement learning has been used to align generative models with external objectives. DDPO~\cite{DDPO} optimizes diffusion sampling with PPO-style updates~\cite{PPO}, Diffusion-DPO~\cite{DiffusionDPO} adapts DPO~\cite{DPO} to pairwise diffusion preferences, and DiffusionNFT~\cite{DiffusionNFT} injects reward signals into the forward flow-matching objective through implicit velocity perturbations, enabling fine-tuning without explicit reverse-trajectory policy-gradient estimation, and has been adopted in several downstream systems~\cite{UniWorld-V2,FireRed-Image-Edit,MeanFlowNFT,Flow-Factory}. Recent Real-ISR alignment works further explore perceptual preferences, LR-conditioned rewards, and fine-grained reward design~\cite{DP2O-SR,RefReward-SR,FinPercep-RM,RealSR-R1}. Unlike these works, LucidNFT targets the multi-reward setting, in which each LR-conditioned rollout group must be ranked by both perceptual realism and LR faithfulness.

Appendix~\ref{sec:supp_related} discusses further Real-ISR alignment and dataset works.  
\vspace{-3mm}
\section{Preliminaries}
\label{sec:prerequisites}
\vspace{-3mm}

\subsection{Flow Matching}
\vspace{-2mm}
Flow matching~\cite{lipman2022flow,liu2022flow} learns a continuous transport map between a source distribution and a target data distribution through a velocity field. Given a data sample $x_0 \sim \mathcal{X}_0$ and a Gaussian sample $x_1 \sim \mathcal{X}_1$, rectified flow constructs a linear interpolation
$
x_t = (1-t)x_0 + t x_1, \quad t \in [0,1].
$
The corresponding velocity field is defined as
$
v = x_1 - x_0.
$  A neural network $v_\theta(x_t,t,c)$ is trained to approximate this velocity by minimizing the flow-matching objective
\begin{equation}
\mathcal{L}_{\mathrm{FM}}(\theta)
=
\mathbb{E}_{t,x_0,x_1}
\left[
\|v - v_\theta(x_t,t,c)\|_2^2
\right].
\end{equation}

For conditional generation tasks such as Real-ISR, the velocity field is conditioned on an observed input $c$, learning a transport map from noise to structured outputs guided by the LR observation.

\vspace{-2mm}
\subsection{Diffusion Negative-aware FineTuning}
\label{sec:prelim_nft}
\vspace{-2mm}
DiffusionNFT~\cite{DiffusionNFT} performs direct policy optimization on the forward diffusion process by leveraging a scalar optimality reward $r(x_0, c)\in[0,1]$. Unlike standard policy-gradient formulations~\cite{Flow-GRPO,DanceGRPO} that optimize reverse-time likelihoods, DiffusionNFT introduces a contrastive diffusion objective that explicitly encourages the velocity predictor to move toward high-reward behaviors and away from low-reward behaviors within the forward flow matching framework. Given the old policy $v_{\text{old}}$ that generates the rollouts and is held fixed within an iteration, DiffusionNFT constructs implicit positive and negative policies:
\begin{align}
v_\theta^+(x_t,t,c) &= (1-\beta_{\mathrm{NFT}})v_{\text{old}}(x_t,t,c) + \beta_{\mathrm{NFT}} v_\theta(x_t,t,c), \\
v_\theta^-(x_t,t,c) &= (1+\beta_{\mathrm{NFT}})v_{\text{old}}(x_t,t,c) - \beta_{\mathrm{NFT}} v_\theta(x_t,t,c),
\end{align}
where the interpolation weight $\beta_{\mathrm{NFT}}$ controls the guidance strength; it is distinct from the coefficient $\lambda_{\mathrm{KL}}$ of a separate KL-style regularizer, the squared distance between the velocity predictions of $v_\theta$ and of the pretrained reference model. These perturbations can be interpreted as shifting the reference velocity field along the improvement direction $v_\theta - v_{\text{old}}$. The training objective is defined as
\begin{equation}
\label{eq:diffnft_obj}
\mathcal{L}(\theta)
=
\mathbb{E}_{c,\,x_0\sim\pi_{\text{old}},\,t}
\left[
r\|v_\theta^+ - v\|_2^2
+
(1-r)\|v_\theta^- - v\|_2^2
\right],
\end{equation}
where $v$ denotes the ground-truth velocity target. High-reward samples emphasize the positive perturbation term, while low-reward samples increase the contribution of the negative term, forming a reward-weighted contrastive update in velocity space. To stabilize optimization, rewards are normalized as
\begin{equation}
r(x_0,c)
=
\frac{1}{2}
+
\frac{1}{2}
\operatorname{clip}
\left(
\frac{r_{\text{raw}}(x_0,c) - \mathbb{E}_{\pi_{\text{old}}}[r_{\text{raw}}]}{Z_c},
-1,1
\right),
\label{eq:diffnft_reward_map}
\end{equation}
where $Z_c$ controls the reward scale and the clipping threshold. In practice, the expectation in Eq.~\eqref{eq:diffnft_reward_map} is estimated from a \emph{rollout group}: several samples are drawn for the same condition $c$, and each reward is compared with the mean of its group, as in group-relative policy optimization~\cite{Flow-GRPO,DanceGRPO}. The resulting advantage is positive for samples that are better than the other samples of the same input, which the update reinforces, and negative for worse ones, which it suppresses; only differences among samples of the same input therefore drive learning. By embedding reward-weighted contrast directly into the forward velocity objective, DiffusionNFT avoids likelihood approximation, supports black-box solvers, and needs only clean samples and their rewards rather than full trajectories. These properties, together with its reported efficiency and downstream adoption (Sec.~\ref{sec:intro}), motivate our choice; in particular, they allow us to retain the backbone's multi-step solver without storing sampling trajectories.
DiffusionNFT, however, consumes a \emph{single} bounded reward $r\in[0,1]$ (Sec.~\ref{sec:multi_reward_rl}).  
\vspace{-4mm}
\section{Method}
\begin{figure*}[t]
    \centering
    \includegraphics[width=\linewidth]{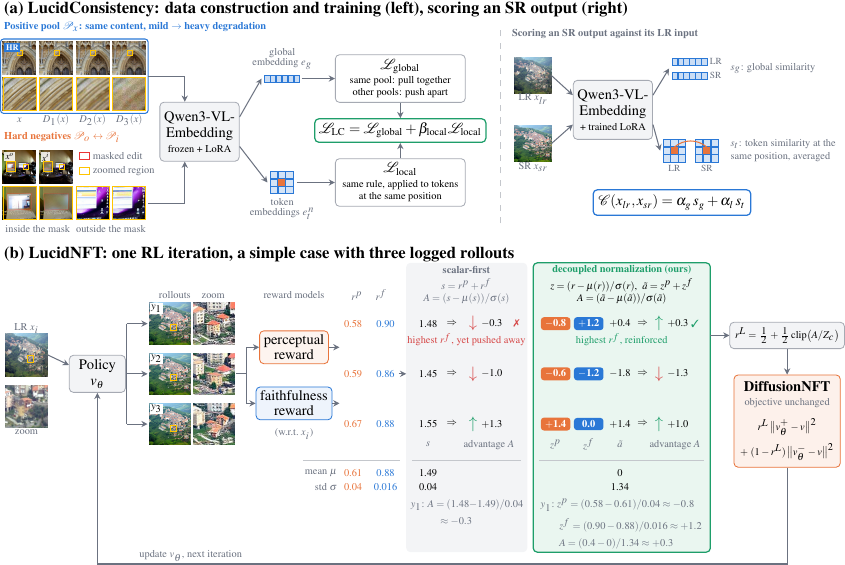}
    \captionsetup{font=scriptsize}
    \vspace{-5mm}
    \caption{
    Overview of LucidNFT.
    (a)~LucidConsistency: the positive pool $\mathcal{P}_x$ holds an HR image $x$ (LSDIR) and its degraded views $D_k(x)$ of increasing severity; an original image $x^{o}$ and its inpainted counterpart $x^{i}$ (SAGI) form pools $\mathcal{P}_o$, $\mathcal{P}_i$ that are hard negatives for each other: they differ inside the masked region and, more subtly, in local details outside it that the regeneration alters (zooms). The global embedding $e_g$ and token embeddings $e_t^{n}$ are trained with $\mathcal{L}_{\mathrm{LC}}$ (Eq.~\eqref{eq:lc_loss}); an SR output $x_{sr}$ is then scored against its LR input $x_{lr}$ by $\mathcal{C}=\alpha_g s_g+\alpha_l s_t$, with global similarity $s_g$ and position-averaged token similarity $s_t$ (Eq.~\eqref{eq:lc_score}).
    (b)~One training iteration of the policy $v_\theta$ on three logged rollouts $y_j$ of one LR input $x_i$ (training uses 24 groups of $M{=}12$). $r^{p}$, $r^{f}$: perceptual and faithfulness rewards; $s$: their sum; $\mu$, $\sigma$: mean and standard deviation over the three rollouts; $z$: a reward standardized within the group; $\tilde a$: the sum of the $z$; $A$: advantage (Eqs.~\eqref{eq:scalar_group_norm}--\eqref{eq:reward_map}), which reinforces an output if positive ($\uparrow$) and pushes it away if negative ($\downarrow$); $r^{L}$: reward weight with scale $Z_c$, which weights the implicit positive and negative policies $v_\theta^{\pm}$ against the target velocity $v$ in DiffusionNFT (Eq.~\eqref{eq:diffnft_obj}). Summing before normalization suppresses the most faithful candidate $y_1$, whereas decoupled normalization reinforces it.
    }
    \label{fig:overview}
    \vspace{-5mm}
\end{figure*}

\vspace{-4mm}
\label{sec:Methodology}
\textbf{LucidNFT} is an LR-anchored multi-reward preference optimization framework for flow-matching Real-ISR built on DiffusionNFT (Sec.~\ref{sec:prelim_nft}): for each LR input from \textbf{LucidLR} (Sec.~\ref{sec:lucidlr}), the policy samples a rollout group, each candidate is scored by a perceptual IQA reward and by \textbf{LucidConsistency} (Sec.~\ref{sec:lucidconsistency}), and \textbf{decoupled reward normalization} (Sec.~\ref{sec:multi_reward_rl}) maps the reward vectors to the reward weight of the unchanged DiffusionNFT objective (Eq.~\eqref{eq:diffnft_obj}); Fig.~\ref{fig:overview} illustrates both stages.

\subsection{LucidConsistency: Degradation-Invariant and Hallucination-Sensitive Consistency Evaluation}
\label{sec:lucidconsistency}
\vspace{-2mm}
Existing Real-ISR evaluation protocols mainly rely on no-reference perceptual metrics~\cite{MUSIQ,CLIP-IQA,Q-Align,NIMA,NIQE,MANIQA}, which assess visual quality but cannot determine whether restored details are supported by the LR observation.
A natural alternative is to compare the LR input and the SR output in a generic semantic embedding space, but such embeddings are not trained for this comparison and agree less with human faithfulness judgments (Table~\ref{tab:human_faithfulness}).
An LR-referenced faithfulness signal for Real-ISR must instead satisfy two requirements at once.
It must be \emph{degradation-invariant}: an LR input and a faithful SR output differ in resolution, blur, noise, and residual artifacts even when their content agrees, and these differences should not be penalized.
It must also be \emph{hallucination-sensitive}: it should still react to the localized, globally plausible changes that generative SR introduces, such as unsupported text strokes, repeated patterns, or fabricated facial parts.
Both requirements are encoded in the training pools: multi-degradation views of an image are positives, and original--inpainted pairs are hard negatives (Fig.~\ref{fig:overview}a). We train LucidConsistency on these pools with standard contrastive learning~\cite{InfoNCE,SupCon}; the design lies in the supervision that the pools define.

\vspace{-3mm}
\paragraph{Degradation invariance: content-consistent pools.}
For an image $x$, we build a pool $\mathcal{P}_{x}=\{x,D_{1}(x),\ldots,D_{K}(x)\}$ of degraded views produced by a randomized Real-ESRGAN pipeline at increasing severity, from mild to heavy; within each level, blur kernel, resizing factor, noise strength, and JPEG quality are sampled at random (ranges in Appendix~\ref{sec:supp_degradation_settings}). Since all views share content and geometry, treating them as positives trains the evaluator to ignore degradation-induced shifts.

\vspace{-3mm}
\paragraph{Hallucination sensitivity: paired hard negatives.}
Invariance to degradation alone would leave the evaluator insensitive to hallucinated content; the training data must therefore also contain negatives that differ from their references only in localized, generator-induced content. Outputs of SR models are ill-suited to this role: their unfaithful regions are not annotated, and training on them would couple the evaluator to the biases of the restoration models that produced them. We therefore draw paired hard negatives from SAGI~\cite{SAGI}, a large-scale dataset of AI-inpainted images. Each pair consists of an original image $x^{o}$ and its inpainted counterpart $x^{i}$; the two are spatially aligned and globally similar but differ within a regenerated region, which mirrors the failure mode of generative SR. We use the fully regenerated subset of SAGI, in which the inpainted images are produced by BrushNet~\cite{BrushNet}, ControlNet inpainting~\cite{ControlNet}, HD-Painter~\cite{HD-Painter}, and PowerPaint~\cite{PowerPaint} on checkpoints of the SD~1.5, SD~2, and SDXL families, as well as by two successive edits that may additionally involve Inpaint Anything~\cite{InpaintAnything}; the hallucinated content thus originates from a diverse set of generators (statistics in Sec.~\ref{sec:impl}). Both images are expanded into degraded pools, $\mathcal{P}_{o}=\{x^{o},D_{1}(x^{o}),\ldots,D_{K}(x^{o})\}$ and $\mathcal{P}_{i}=\{x^{i},D_{1}(x^{i}),\ldots,D_{K}(x^{i})\}$, which serve as hard negatives for each other. As the degradation pipeline and the inpainting generators are unrelated, invariance and sensitivity are learned from independent sources of variation; Sec.~\ref{sec:lc_eval} examines robustness to the choice of inpainting generators with a leave-one-inpainter-out protocol.

\vspace{-3mm}
\paragraph{Global-local representation.}
We use Qwen3-VL-Embedding~\cite{Qwen3-VL-Embedding} as the backbone and train only LoRA adapters. For each image view $v$, we extract a global embedding $e_g(v)$ and native image-token embeddings $E_t(v)=\{e_t^i(v)\}_{i=1}^{N_v}$, where $N_v$ is the number of image tokens produced for $v$. The global embedding captures image-level semantic consistency, while native tokens provide spatially indexed local correspondence without manually defining crops or windows. All views are resized to a shared spatial size before encoding, so $N_v=N$ within a batch and same-index tokens correspond to the same backbone-defined visual positions.

\vspace{-3mm}
\paragraph{Pool-based contrastive learning.}
For each query representation $q$, positives are representations from other views in the same pool, while negatives are representations from other pools. Let $\mathcal{Q}$ be the set of valid queries in a mini-batch, $\mathcal{P}^{+}(q)$ the same-pool positive set excluding $q$, and $\mathcal{P}^{-}(q)$ the negative set from other pools. We use
\begin{equation}
\mathcal{L}_{\mathrm{ctr}}
=
-
\frac{1}{|\mathcal{Q}|}
\sum_{q\in\mathcal{Q}}
\log
\frac{
\sum\limits_{p\in\mathcal{P}^{+}(q)}
\exp\left(\mathrm{s}(q,p)/\tau\right)
}{
\sum\limits_{p\in\mathcal{P}^{+}(q)}
\exp\left(\mathrm{s}(q,p)/\tau\right)
+
\sum\limits_{n\in\mathcal{P}^{-}(q)}
\exp\left(\mathrm{s}(q,n)/\tau\right)
}.
\label{eq:pool_contrastive}
\end{equation}
For the global branch, $q,p,n$ are global embeddings. For the local branch, $q$ is a token embedding $e_t^i(\cdot)$, positives are same-index tokens from same-pool views, and negatives are same-index tokens from other pools. Applying Eq.~\eqref{eq:pool_contrastive} to the two branches gives
\begin{equation}
\mathcal{L}_{LC}
=
\mathcal{L}_{global}
+
\beta_{local}\mathcal{L}_{local}.
\label{eq:lc_loss}
\end{equation}

\vspace{-3mm}
\paragraph{Inference score.}
Given an LR input $x_{lr}$ and a generated SR output $x_{sr}$, LucidConsistency combines global and local consistency:
\begin{equation}
\mathcal{C}(x_{lr},x_{sr})
=
\alpha_g\,\mathrm{s}\big(e_g(x_{lr}),e_g(x_{sr})\big)
+
\alpha_l\,\frac{1}{N}
\sum_{i=1}^{N}
\mathrm{s}\big(e_t^i(x_{lr}),e_t^i(x_{sr})\big),
\qquad
\alpha_g+\alpha_l=1.
\label{eq:lc_score}
\end{equation}
In all experiments, we set $\tau=0.7$, $\beta_{\mathrm{local}}=1.0$, and use equal global-local weights $\alpha_g=\alpha_l=0.5$ unless otherwise specified. This score serves as both an evaluation metric and an RL reward.

\vspace{-2mm}
\subsection{Decoupled Reward Normalization for Rollout-Group Real-ISR}
\label{sec:multi_reward_rl}
\vspace{-2mm}

For a fixed LR input, stochastic sampling produces a rollout group of candidate restorations that differ along the perceptual--faithfulness trade-off: a sharper output may hallucinate unsupported details, while a more conservative one may better preserve the LR evidence.
The useful supervision is therefore the reward contrast among candidates of the same LR input, rather than the absolute reward of each sample.
DiffusionNFT, however, consumes this contrast through a single bounded reward weight $r\in[0,1]$ (Sec.~\ref{sec:prelim_nft}), whereas Real-ISR must balance at least two heterogeneous rewards.
The bottleneck we address lies at this scalar-reward interface: how a multi-dimensional reward vector is mapped to $r$ decides whether both contrasts survive.

\noindent \textit{\textbf{Notation}.}
Let $x_i$ denote the $i$-th LR input and $y_{i,j}$ its $j$-th rollout within the same LR-conditioned group.
For each rollout, we compute a $K$-dimensional reward vector $\mathbf{r}_{i,j}=(r_{i,j}^{(1)},\ldots,r_{i,j}^{(K)})$, where each dimension corresponds to one objective and has weight $\lambda_k$. We write $\mu_i,\sigma_i$ for the mean and standard deviation within the group of $x_i$, and $\mu_B,\sigma_B$ for those over the training batch.

\noindent \textit{\textbf{Scalar-first aggregation}.}
A common strategy first fuses rewards into a scalar score $s_{i,j}=\sum_{k=1}^{K}\lambda_k r_{i,j}^{(k)}$, and then normalizes the scalar scores within rollout groups:
\begin{equation}
A_{i,j}
=
\frac{s_{i,j}-\mu_i(s)}{\sigma_B(s)+\epsilon},
\label{eq:scalar_group_norm}
\end{equation}
where $\mu_i(s)=\frac{1}{M_i}\sum_{j=1}^{M_i}s_{i,j}$.

\begin{figure*}[t]
    \centering
    \vspace{-2mm}
    \includegraphics[width=\linewidth]{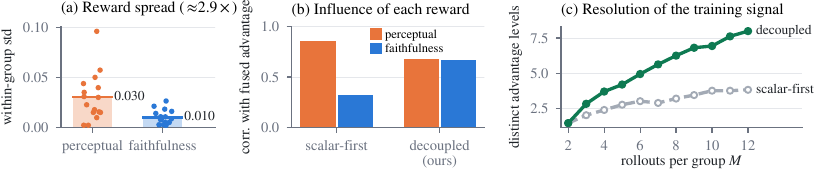}
    \captionsetup{font=scriptsize}
    \vspace{-5mm}
    \caption{Failure mode of scalar-first fusion (16 LR inputs $\times$ 12 LucidFlux rollouts). (a)~The perceptual reward varies about $3\times$ more within a group than the faithfulness reward. (b)~Summing before normalization therefore yields advantages that follow the perceptual reward, whereas decoupled normalization gives both rewards equal influence (within-group Pearson correlation). (c)~It also separates more candidates.}
    \label{fig:mechanism}
    \vspace{-1mm}
\end{figure*}

\noindent \textit{\textbf{Failure mode of scalar-first fusion}.}
Consider a group of three candidates with (perceptual, faithfulness) rewards $(1,0)$, $(0.5,0.5)$, and $(0,1)$. With $\lambda_1=\lambda_2=1$, all three obtain $s_{i,j}=1$ and therefore $A_{i,j}=0$: the group contains two opposite orderings, yet the update receives no preference signal. Normalizing each reward first yields $z^{(1)}=(+1.22,0,-1.22)$ and $z^{(2)}=(-1.22,0,+1.22)$, which preserves both orderings; with equal weights this symmetric case still sums to zero.
Real rollout groups are rarely this symmetric; in practice, scalar-first fusion instead reweights the objectives implicitly. On 16 held-out LR inputs with 12 LucidFlux rollouts each, both rewards lie in $[0,1]$ and carry nominal weights $1{:}1$, yet their within-group standard deviations differ by a factor of $0.0305/0.0104\approx2.94$ (Fig.~\ref{fig:mechanism}a). Under scalar-first fusion, the fused advantage correlates with the perceptual reward at $0.861$ but with the faithfulness reward at only $0.318$, so the effective weighting is closer to $2.7{:}1$ than to the specified $1{:}1$ (Fig.~\ref{fig:mechanism}b). Fig.~\ref{fig:overview}b applies both fusion rules to the logged rewards of three rollouts of one training input: scalar-first fusion would suppress the most faithful candidate, whereas decoupled normalization reinforces it.
As DiffusionNFT maps advantages to a bounded weight $r\in[0,1]$, the under-weighted objective also occupies little of the range around $r=0.5$. We refer to this as \emph{advantage compression}.

\noindent \textit{\textbf{Decoupled normalization before fusion}.}
LucidNFT preserves objective-wise contrasts by normalizing each reward dimension within the same LR-conditioned rollout group before fusion:
\begin{equation}
z_{i,j}^{(k)}
=
\frac{
r_{i,j}^{(k)}-\mu_i(r^{(k)})
}{
\sigma_i(r^{(k)})+\epsilon
},
\label{eq:dim_norm}
\end{equation}
We then fuse the normalized rewards as $\tilde{a}_{i,j}=\sum_{k=1}^{K}\lambda_k z_{i,j}^{(k)}$.
Finally, we standardize $\tilde a$ over the batch and map the result to the bounded DiffusionNFT reward weight:
\begin{equation}
A_{i,j}^{L}
=
\frac{
\tilde{a}_{i,j}-\mu_B(\tilde a)
}{
\sigma_B(\tilde a)+\epsilon
},
\qquad
r_{i,j}^{L}
=
\frac{1}{2}
+
\frac{1}{2}
\operatorname{clip}
\left(
\frac{A_{i,j}^{L}}{Z_c},
-1,1
\right).
\label{eq:reward_map}
\end{equation}
Since each $z^{(k)}$ is standardized within its group, $\mu_B(\tilde a)=0$ and this step only rescales $\tilde a$. We substitute $r_{i,j}^{L}$ for the reward weight in the DiffusionNFT objective and keep all other terms unchanged.
On the same rollouts, the correlations of the fused advantage with the perceptual and faithfulness rewards become $0.680$ and $0.674$, so the two objectives contribute almost equally, as the $1{:}1$ weights specify (Fig.~\ref{fig:mechanism}b).

\vspace{-1mm}
\noindent \textit{\textbf{Resolution of the training signal}.}
Preserving both orderings also makes the advantages more informative. Fig.~\ref{fig:mechanism}c counts the distinct advantage levels within a group on the same rollouts: with 12 rollouts, decoupled normalization separates the candidates into 8.0 levels on average, whereas scalar-first fusion yields 3.8, leaving many candidates with different trade-offs at the same advantage. The advantage gap between the two most different candidates grows accordingly, from 0.72 to 2.19 at $M{=}12$ (Appendix~\ref{sec:supp_adv}).

\begin{figure*}[ht]
    \vspace{1.5mm}
    \centering
    
    \begin{minipage}{0.42\textwidth}
        \centering
        \includegraphics[width=\linewidth]{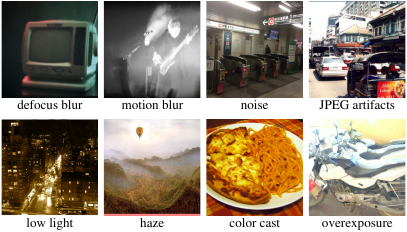}
        \vspace{-6mm}
        \captionsetup{font=scriptsize} 
        \captionof{figure}{LucidLR inputs for Real-ISR: real low-quality photographs, each labeled with the dominant degradation that accompanies its low resolution.}
        \label{fig:lucidlr}
    \end{minipage}
    \hfill
    \begin{minipage}{0.5\textwidth}
        \centering
        \captionsetup{font=scriptsize} 
        \captionof{table}{Real-world Real-ISR datasets. Degradation statistics use the same annotator and taxonomy for all datasets (Appendix~\ref{sec:supp_lucidlr_stats}): number of primary-degradation types observed (excluding \emph{other} and \emph{none}), and share of images whose primary degradation is defocus blur.}
        \vspace{-2mm}
        \label{tab:dataset_compare}
        \setlength{\tabcolsep}{5pt}
        \renewcommand{\arraystretch}{1.15}
        \resizebox{\linewidth}{!}{
            \begin{tabular}{l c c r c c}
                \toprule
                \textbf{Dataset} & \textbf{Pairing} & \textbf{Usage} & \textbf{\# Images} & \textbf{Degr.\ types} & \textbf{Defocus (\%)} \\
                \midrule
                RealSR~\cite{RealSR}        & Paired   & Benchmark   & 100    & 4  & 82.0 \\
                DRealSR~\cite{DRealSR}      & Paired   & Benchmark   & 93     & 5  & 91.4 \\
                RealLQ250~\cite{DreamClear} & Unpaired & Benchmark   & 250    & 10 & 43.2 \\
                \midrule
                \textbf{LucidLR (ours)}     & Unpaired & RL training & 20,000 & 15 & 26.5 \\
                \bottomrule
            \end{tabular}
        }
    \end{minipage}
    
    \vspace{-6mm}
\end{figure*}

\subsection{LucidLR: A Large-Scale Real-World Degradation Dataset for Real-ISR}
\label{sec:lucidlr}
\vspace{-2mm}
Existing real-world Real-ISR datasets~\cite{RealSR,DRealSR,DreamClear} are small benchmarks (Table~\ref{tab:dataset_compare}), whereas RL fine-tuning benefits from a large and diverse set of real degradations that induce informative rollout variation. We therefore construct \textbf{LucidLR} from the Wikimedia Commons categories \emph{Images of low quality} and \emph{Blurred images}, whose photographs exhibit naturally occurring degradations and may be reused with attribution. The raw pool contains 21.7K unique images. A two-pass screening with a vision-language model (DeepSeek-V4.1-Flash~\cite{DeepSeekV4}) removes 72 images containing explicit sexual content, graphic violence, personal data, or hate symbols; we further exclude images whose short side is below 128 pixels or whose source pages are no longer available, and supplement the pool with images collected under the same protocol, yielding \textbf{20,000} images (Fig.~\ref{fig:lucidlr}). LucidLR contains only images whose licenses permit redistribution. It is released both as an image bundle and as a metadata index with a reference downloader, and each image is accompanied by a JSON sidecar that records its source, author, and license (Appendix~\ref{sec:supp_lucidlr_release}). Annotated with the same model and taxonomy as existing benchmarks, LucidLR spans all 13 semantic categories and 15 primary-degradation types, and far fewer of its images are dominated by defocus blur (Table~\ref{tab:dataset_compare}); its images are not resized and vary widely in resolution (median $1080\times960$; Appendix~\ref{sec:supp_lucidlr_stats}).  
\vspace{-4mm}
\section{Experiments}
\vspace{-3mm}
\subsection{Implementation Details}
\label{sec:impl}
\vspace{-1mm}

For \textbf{LucidConsistency}, we adopt Qwen3-VL-Embedding-8B~\cite{Qwen3-VL-Embedding} as the backbone and train LoRA adapters while keeping the backbone frozen. The training data consist of two parts. First, 85K LSDIR~\cite{LSDIR} images are used to build content-consistent pools for degradation-invariant alignment, where multi-level views are generated by the Real-ESRGAN~\cite{Real-ESRGAN} degradation pipeline. We use three ordered degradation levels (mild, medium, and heavy), with detailed parameter ranges provided in Appendix~\ref{sec:supp_degradation_settings}. Second, original--inpainted hard negatives come from SAGI~\cite{SAGI}: we use all fully regenerated pairs of its training split whose originals are from COCO~\cite{COCO} and RAISE~\cite{RAISE}, 37.7K pairs from 33.3K originals. Of these, 27.8K pairs are edited once, by one of 11 method--checkpoint combinations that SAGI provides, i.e., each inpainting method on the checkpoints it supports (Fig.~\ref{fig:overview}a): BrushNet~\cite{BrushNet} on SD~1.5~\cite{SD}, DreamShaper, Realistic Vision, epiCRealism, SDXL~\cite{SDXL}, and Juggernaut XL; ControlNet inpainting~\cite{ControlNet} on SD~1.5; HD-Painter~\cite{HD-Painter} on SD~1.5, SD~2, and DreamShaper; and PowerPaint~\cite{PowerPaint} on Realistic Vision. These seven checkpoints belong to three base families: SD~1.5 and its fine-tunes, SD~2, and SDXL with its fine-tune Juggernaut XL; the remaining 9.9K pairs are edited twice in succession, each round by one of these four methods or by Inpaint Anything~\cite{InpaintAnything} on SD~2 (2.9K pairs). Every inpainted variant forms its own pool, a hard negative for its original and an ordinary negative for all other pools; each pair is brought to a common pixel grid and cropped around the edit (Appendix~\ref{sec:supp_degradation_settings}). We train LucidConsistency at resolution 1024 with LoRA rank and alpha both 128, using batch size 2 on 2 NVIDIA A100 GPUs for 6,000 steps. The learning rate is $1\times10^{-4}$. For \textbf{LucidNFT}, we fine-tune the base Real-ISR model using LoRA~\cite{LoRA} (rank/alpha 32/64) on the LucidLR dataset. Training runs on 8 NVIDIA A100 GPUs for 600 RL steps with a learning rate of $3\times10^{-5}$, using 24 LR groups of 12 rollouts per step at resolution 768; solver, $\beta_{\mathrm{NFT}}$, $\lambda_{\mathrm{KL}}$, EMA, and the remaining hyperparameters are listed with pseudocode in Appendix~\ref{sec:supp_impl}. We use UniPercept IQA~\cite{UniPercept} as the perceptual reward and LucidConsistency as the LR-faithfulness reward, both with weight 1.0. Both models are optimized with AdamW~\cite{AdamW}.

\vspace{-4mm}
\subsection{Human-Aligned Faithfulness Evaluation}
\label{sec:lc_eval}
\vspace{-2mm}
We further evaluate whether LucidConsistency reflects human judgments of LR-conditioned faithfulness.
On RealSR~\cite{RealSR}, we collect SR candidates from eight generative SR methods: ResShift~\cite{Resshift}, SeeSR~\cite{SeeSR}, LucidFlux~\cite{LucidFlux}, DreamClear~\cite{DreamClear}, SUPIR~\cite{SUPIR}, DiffBIRv2~\cite{DiffBIR}, DiT4SR~\cite{DiT4SR}, and StableSR~\cite{StableSR}.
For each LR input, five expert annotators rank four randomly sampled candidates by faithfulness to the LR evidence rather than sharpness or aesthetics, yielding six pairwise comparisons (instructions and interface in Appendix~\ref{sec:supp_human_pref_lr_faithfulness}).

We compare evaluator-induced rankings with human preferences using pairwise Agreement, Recall@1 (selecting the human-preferred best candidate), and Filter@1 (assigning the lowest score to the human-labeled worst candidate).
As shown in Table~\ref{tab:human_faithfulness}, LucidConsistency achieves the best results on all three metrics.
Compared with the frozen semantic baseline, it improves Agreement from 0.643 to 0.690, Recall@1 from 0.465 to 0.558, and Filter@1 from 0.302 to 0.558, indicating that pool-based training better calibrates generic semantic representations toward LR-conditioned faithfulness.
Perceptual metrics perform substantially worse under this protocol, suggesting that visual quality alone is insufficient for identifying hallucinated or LR-inconsistent details.

\noindent \textbf{Robustness to the hard-negative construction.}
To test whether LucidConsistency overfits to its synthetic inpainting negatives, we retrain it without every pair in which an SDXL-family generator (BrushNet with SDXL or Juggernaut XL) takes part in either editing round (leave-one-inpainter-out, LOIO), thereby withholding an entire base architecture.
On the SAGI test split of the held-out generators and on the 40 human-labeled RealSR pairs of Sec.~\ref{sec:effectiveness}, the paired agreement drops only from 0.74 and 0.69 (full) to 0.70 and 0.67 (LOIO), both well above chance, which suggests that the learned faithfulness signal does not hinge on the generators seen in training (Appendix~\ref{sec:supp_loio}).

\vspace{-4mm}
\begin{table*}[t]
\centering
\begin{minipage}[t]{0.48\textwidth}
\centering
\captionsetup{font=scriptsize}
\captionof{table}{
Human-aligned LR-faithfulness evaluation on RealSR. 
Expert annotators rank SR candidates by faithfulness to the LR input rather than perceptual quality.
}
\vspace{-2mm}
\setlength{\tabcolsep}{5.0pt}
\renewcommand{\arraystretch}{1.28}
\begin{adjustbox}{max width=\linewidth}
\begin{tabular}{l l c c c}
\toprule
\textbf{Evaluator} & \textbf{Criterion} & \textbf{Agreement} & \textbf{Recall@1} & \textbf{Filter@1} \\
\midrule
CLIP-IQA & Perceptual Quality & 0.391 & 0.186 & 0.093 \\
MUSIQ & Perceptual Quality & 0.349 & 0.116 & 0.093 \\
Q-Align & Perceptual Quality & 0.322 & 0.093 & 0.047 \\
UniPercept-IQA & Perceptual Quality & 0.322 & 0.070 & 0.093 \\
\midrule
Qwen3-VL-Embedding-8B & Generic Semantics & 0.643 & 0.465 & 0.302 \\
LucidConsistency & LR Faithfulness & \textbf{0.690} & \textbf{0.558} & \textbf{0.558} \\
\bottomrule
\end{tabular}
\end{adjustbox}
\label{tab:human_faithfulness}
\end{minipage}
\hfill
\begin{minipage}[t]{0.48\textwidth}
\centering
\captionsetup{font=scriptsize}
\captionof{table}{Controlled ablation on RealLQ250 (LucidFlux). First row: absolute values; other rows: changes relative to it (larger is better). Each row modifies only the stated variable.}
\vspace{-2mm}
\setlength{\tabcolsep}{3.0pt}
\renewcommand{\arraystretch}{1.10}
\begin{adjustbox}{max width=\linewidth}
\begin{tabular}{lcccc}
\toprule
\textbf{Method} & \textbf{UniPercept} $\uparrow$ & \textbf{VQ-R1} $\uparrow$ & \textbf{NIQE} $\downarrow$ & \textbf{LucidConsistency} $\uparrow$ \\
\midrule
LucidFlux (backbone) & 70.930 & 4.547 & 3.741 & 0.9237 \\
\midrule
\multicolumn{5}{l}{\textit{DiffusionNFT baselines}} \\
(A) UniPercept IQA reward & +0.608 & +0.054 & $-$0.227 & +0.0022 \\
(B) (A) + frozen Qwen3-VL reward & +0.436 & +0.049 & $-$0.212 & +0.0061 \\
\midrule
\multicolumn{5}{l}{\textit{Our components}} \\
(C) (B), semantic $\to$ LucidConsistency & +0.284 & +0.046 & $-$0.194 & +0.0104 \\
(D) (C), LSDIR $\to$ LucidLR & +1.40 & +0.062 & $-$0.345 & +0.0114 \\
(E) (C), scalar-first $\to$ decoupled & +1.773 & +0.086 & $-$0.369 & \textbf{+0.0119} \\
(F) (E), LSDIR $\to$ LucidLR (Full) & \textbf{+2.549} & \textbf{+0.104} & \textbf{$-$0.488} & +0.0108 \\
\bottomrule
\end{tabular}
\end{adjustbox}
\label{tab:ablation}
\end{minipage}
\vspace{-8mm}
\end{table*}

\subsection{Effectiveness of LucidNFT}
\label{sec:effectiveness}
\vspace{-2mm}

We evaluate LucidNFT on two representative flow-based Real-ISR models, LucidFlux~\cite{LucidFlux} and DiT4SR~\cite{DiT4SR}. 
LucidFlux is a 14B DiT-based model with a 12B flow prior, whereas DiT4SR is a lighter DiT derived from Stable-Diffusion-3-Medium~\cite{SD3} (2B prior and 2.72B adapter), allowing us to test LucidNFT across backbone scales.

During training, the validation UniPercept IQA reward of LucidNFT rises steadily on both backbones while LucidConsistency stays stable (Appendix Fig.~\ref{fig:curve_lucidflux}).

\vspace{-1mm}
\noindent \textbf{Comparison Methods and Evaluation Metrics.} 
We compare LucidFlux(+LucidNFT) with representative generative SR methods, including DiffBIRv2~\cite{DiffBIR}, SeeSR~\cite{SeeSR}, DreamClear~\cite{DreamClear}, SUPIR~\cite{SUPIR}, DiT4SR~\cite{DiT4SR}, and LucidFlux~\cite{LucidFlux}. 
We also include DiT4SR(+LucidNFT) and LucidFlux(+DPO), implemented with Diffusion-DPO~\cite{DiffusionDPO}, to compare against direct preference optimization under related backbones. 
We use this same-backbone DPO implementation as a controlled direct-preference baseline; DP$^2$O-SR~\cite{DP2O-SR} is closely related but its public release does not include training code, as discussed in Appendix~\ref{sec:supp_related}.
All methods are evaluated at $1024\times1024$ output resolution with $4\times$ upscaling, following official inference settings when available. 
We report eight no-reference quality metrics, including CLIP-IQA+~\cite{CLIP-IQA}, Q-Align~\cite{Q-Align}, MUSIQ~\cite{MUSIQ}, MANIQA~\cite{MANIQA}, CLIP-IQA~\cite{CLIP-IQA}, NIQE~\cite{NIQE}, UniPercept IQA~\cite{UniPercept}, and VisualQuality-R1 (VQ-R1)~\cite{VisualQuality-R1}. 
We additionally report LucidConsistency as an LR-referenced consistency score without HR ground truth.

\begin{table}[t]
\centering
\captionsetup{font=scriptsize}
\caption{
Quantitative comparison with state-of-the-art Real-ISR methods on RealLQ250~\cite{DreamClear}, DRealSR~\cite{DRealSR}, and RealSR~\cite{RealSR}. 
Higher is better for all metrics except NIQE. 
Values in parentheses are changes over the corresponding backbone. 
Best and second-best scores are highlighted in bold and underline, respectively. 
For the parenthetical changes, $^\dagger$ and $^\ddagger$ mark the largest and second-largest improvements among the RL variants; for NIQE, larger improvement means a larger reduction.
}
\label{tab:quantitative}

\setlength{\tabcolsep}{3.4pt}
\renewcommand{\arraystretch}{1.1}

\begin{adjustbox}{max width=0.98\linewidth}
\begin{tabular}{l l c c c c c c c c c}
\toprule
\textbf{Benchmark} & \textbf{Metric} & \textbf{DiffBIRv2} & \textbf{SeeSR} & \textbf{DreamClear} & \textbf{SUPIR} & \textbf{DiT4SR} & \textbf{DiT4SR(+LucidNFT)} & \textbf{LucidFlux} & \textbf{LucidFlux(+DPO)} & \textbf{LucidFlux(+LucidNFT)} \\
\midrule

\multirow{9}{*}{\textbf{RealLQ250}}
& CLIP-IQA+ $\uparrow$ & 0.6919 & 0.7034 & 0.6813 & 0.6532 & 0.7098 & 0.7124 \secondgain{+0.0026} & 0.7208 & \underline{0.7228} \gain{+0.0020} & \textbf{0.7465} \bestgain{+0.0257} \\
& Q-Align $\uparrow$ & 3.9755 & 4.1423 & 4.0647 & 4.1347 & 4.2270 & 4.2358 \gain{+0.0088} & 4.4052 & \underline{4.4430} \secondgain{+0.0378} & \textbf{4.4855} \bestgain{+0.0803} \\
& MUSIQ $\uparrow$ & 67.5313 & 70.3757 & 67.0899 & 65.8133 & 71.6682 & 72.1732 \secondgain{+0.5050} & 72.3351 & \underline{72.4504} \gain{+0.1153} & \textbf{73.4475} \bestgain{+1.1124} \\
& MANIQA $\uparrow$ & 0.4900 & 0.4895 & 0.4405 & 0.3826 & 0.4607 & 0.4719 \secondgain{+0.0112} & 0.5227 & \underline{0.5258} \gain{+0.0031} & \textbf{0.5443} \bestgain{+0.0216} \\
& CLIP-IQA $\uparrow$ & 0.7137 & 0.7063 & 0.6957 & 0.5767 & 0.7141 & \textbf{0.7355} \secondgain{+0.0214} & 0.6855 & 0.6917 \gain{+0.0062} & \underline{0.7233} \bestgain{+0.0378} \\
& NIQE $\downarrow$ & 5.1193 & 4.4383 & 3.8709 & 3.6591 & 3.5556 & \underline{3.5007} \secondgain{-0.0549} & 3.7410 & 3.7785 \gain{+0.0375} & \textbf{3.2532} \bestgain{-0.4878} \\
& UniPercept IQA $\uparrow$ & 65.4760 & 69.2015 & 68.8465 & 68.6430 & 73.0740 & \underline{73.3430} \secondgain{+0.2690} & 70.9300 & 71.1330 \gain{+0.2030} & \textbf{73.4790} \bestgain{+2.5490} \\
& VisualQuality-R1 $\uparrow$ & 4.3428 & 4.5118 & 4.4430 & 4.4265 & 4.6146 & \underline{4.6304} \gain{+0.0158} & 4.5474 & 4.5644 \secondgain{+0.0170} & \textbf{4.6510} \bestgain{+0.1036} \\
& LucidConsistency $\uparrow$ & \textbf{0.9609} & 0.9466 & \underline{0.9578} & 0.9522 & 0.9052 & 0.9172 \bestgain{+0.0120} & 0.9237 & 0.9296 \gain{+0.0059} & 0.9345 \secondgain{+0.0108} \\

\cmidrule(lr){1-11}

\multirow{9}{*}{\textbf{DRealSR}}
& CLIP-IQA+ $\uparrow$ & 0.6476 & 0.6258 & 0.4462 & 0.5494 & 0.6537 & \underline{0.6757} \secondgain{+0.0220} & 0.6516 & 0.6530 \gain{+0.0014} & \textbf{0.6867} \bestgain{+0.0351} \\
& Q-Align $\uparrow$ & 3.0487 & 3.2746 & 2.4214 & 3.4722 & 3.6008 & 3.6641 \secondgain{+0.0633} & 3.7141 & \underline{3.7408} \gain{+0.0267} & \textbf{3.8423} \bestgain{+0.1282} \\
& MUSIQ $\uparrow$ & 60.0759 & 61.3222 & 35.1912 & 54.9280 & 63.8051 & \underline{65.1915} \secondgain{+1.3864} & 64.6025 & 64.5607 \gain{-0.0418} & \textbf{68.1545} \bestgain{+3.5520} \\
& MANIQA $\uparrow$ & \underline{0.4900} & 0.4505 & 0.2676 & 0.3483 & 0.4419 & 0.4572 \secondgain{+0.0153} & 0.4678 & 0.4669 \gain{-0.0009} & \textbf{0.5004} \bestgain{+0.0326} \\
& CLIP-IQA $\uparrow$ & 0.6782 & 0.6760 & 0.4361 & 0.5310 & 0.6732 & \textbf{0.7111} \secondgain{+0.0379} & 0.6673 & 0.6713 \gain{+0.0040} & \underline{0.7073} \bestgain{+0.0400} \\
& NIQE $\downarrow$ & 6.4853 & 6.4503 & 7.0164 & 5.9092 & 5.7001 & 5.6329 \secondgain{-0.0672} & 5.0742 & \underline{5.0143} \gain{-0.0599} & \textbf{4.1788} \bestgain{-0.8954} \\
& UniPercept IQA $\uparrow$ & 46.2298 & 50.3414 & 34.2473 & 55.1371 & 58.1290 & \underline{59.9328} \secondgain{+1.8038} & 59.9032 & 59.7782 \gain{-0.1250} & \textbf{63.7782} \bestgain{+3.8750} \\
& VisualQuality-R1 $\uparrow$ & 3.4796 & 3.6116 & 2.5655 & 3.7349 & 3.9603 & \underline{4.0239} \secondgain{+0.0636} & 3.9955 & 3.9828 \gain{-0.0127} & \textbf{4.1455} \bestgain{+0.1500} \\
& LucidConsistency $\uparrow$ & \underline{0.9332} & 0.9275 & \textbf{0.9607} & 0.8911 & 0.8438 & 0.8544 \bestgain{+0.0106} & 0.8813 & 0.8890 \secondgain{+0.0077} & 0.8879 \gain{+0.0066} \\

\cmidrule(lr){1-11}

\multirow{9}{*}{\textbf{RealSR}}
& CLIP-IQA+ $\uparrow$ & 0.6543 & 0.6731 & 0.5331 & 0.5640 & 0.6753 & \underline{0.6881} \secondgain{+0.0128} & 0.6669 & 0.6695 \gain{+0.0026} & \textbf{0.7151} \bestgain{+0.0482} \\
& Q-Align $\uparrow$ & 3.3156 & 3.6073 & 3.0040 & 3.4682 & 3.7106 & 3.7959 \secondgain{+0.0853} & 3.8728 & \underline{3.9147} \gain{+0.0419} & \textbf{3.9918} \bestgain{+0.1190} \\
& MUSIQ $\uparrow$ & 61.7751 & 67.5660 & 49.4766 & 55.6807 & 67.9828 & \underline{69.1092} \secondgain{+1.1264} & 67.8962 & 67.9362 \gain{+0.0400} & \textbf{70.5625} \bestgain{+2.6663} \\
& MANIQA $\uparrow$ & 0.4745 & \underline{0.5087} & 0.3092 & 0.3426 & 0.4533 & 0.4654 \secondgain{+0.0121} & 0.4889 & 0.4907 \gain{+0.0018} & \textbf{0.5284} \bestgain{+0.0395} \\
& CLIP-IQA $\uparrow$ & 0.6806 & \textbf{0.6993} & 0.5390 & 0.4857 & 0.6631 & \underline{0.6963} \secondgain{+0.0332} & 0.6359 & 0.6427 \gain{+0.0068} & 0.6936 \bestgain{+0.0577} \\
& NIQE $\downarrow$ & 6.0700 & 5.4594 & 5.2873 & 5.2819 & 5.0912 & 4.8332 \secondgain{-0.2580} & 4.8134 & \underline{4.6804} \gain{-0.1330} & \textbf{3.9526} \bestgain{-0.8608} \\
& UniPercept IQA $\uparrow$ & 53.6550 & 58.0538 & 46.7850 & 56.6063 & 63.2025 & \textbf{64.8425} \secondgain{+1.6400} & 60.0925 & 60.4775 \gain{+0.3850} & \underline{64.7588} \bestgain{+4.6663} \\
& VisualQuality-R1 $\uparrow$ & 3.8928 & 4.0635 & 3.5028 & 3.7821 & 4.1953 & \underline{4.2429} \secondgain{+0.0476} & 4.1376 & 4.1503 \gain{+0.0127} & \textbf{4.3389} \bestgain{+0.2013} \\
& LucidConsistency $\uparrow$ & \textbf{0.9544} & 0.9138 & \underline{0.9475} & 0.9141 & 0.8318 & 0.8498 \bestgain{+0.0180} & 0.8853 & 0.8932 \secondgain{+0.0079} & 0.8923 \gain{+0.0070} \\

\bottomrule
\end{tabular}
\end{adjustbox}
\vspace{-8mm}
\end{table}

\vspace{-1mm}
\noindent \textbf{Quantitative Comparison.}
Table~\ref{tab:quantitative} reports quantitative results on RealLQ250~\cite{DreamClear}, DRealSR~\cite{DRealSR}, and RealSR~\cite{RealSR}. 
Both DiT4SR(+LucidNFT) and LucidFlux(+LucidNFT) improve their corresponding backbones on most perceptual metrics while generally maintaining or improving LucidConsistency.
On LucidFlux, LucidNFT improves all eight no-reference metrics on all three benchmarks, with the largest relative gains on NIQE (reductions of 13.0\%, 17.7\%, and 17.9\% on RealLQ250, DRealSR, and RealSR) and MANIQA.
LucidFlux(+DPO), which shares the backbone and training data, provides a comparison with direct preference optimization: averaged over the three benchmarks, LucidNFT improves the eight no-reference metrics by 2.8\% (Q-Align) to 16.2\% (NIQE reduction) and LucidConsistency by 0.9\%, whereas Diffusion-DPO changes the perceptual metrics by no more than about 1\% on average and degrades 5 of the 24 benchmark--metric entries, all of which LucidNFT improves (Appendix~\ref{sec:supp_relative}).

\vspace{-1mm}
\noindent \textbf{Model-level human evaluation.}
As LucidConsistency also serves as a training reward, its gains in Table~\ref{tab:quantitative} are not fully external evidence of faithfulness.
We therefore run a human study without LucidConsistency scores: on 40 held-out RealSR inputs, five expert annotators see the LR image and the outputs of LucidFlux and LucidFlux(+LucidNFT) in random order, and select the output better supported by the LR content, disregarding sharpness, saturation, and aesthetics.
Over 200 binary judgments, LucidFlux(+LucidNFT) is preferred in 72.5\% (Fleiss' $\kappa=0.415$, moderate agreement~\cite{LandisKoch}; Appendix~\ref{sec:supp_model_human}).

\begin{figure*}[ht]
    \centering
    \includegraphics[page=1, width=0.92\linewidth]{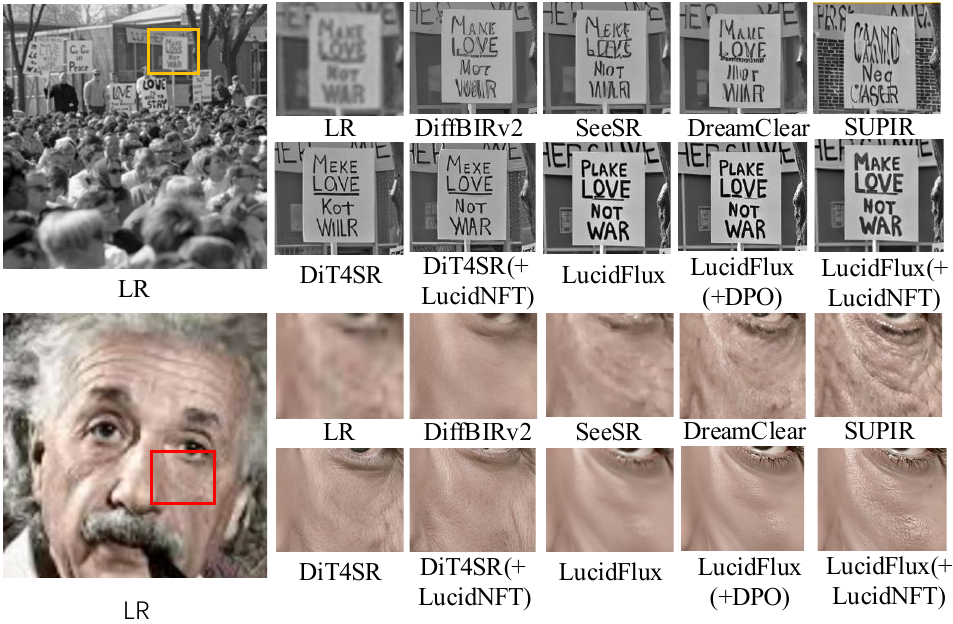}
    \vspace{-2mm}
    \captionsetup{font=scriptsize} 
    \caption{Visual comparison on RealLQ250~\cite{DreamClear}; LucidNFT variants recover more accurate text and finer textures than the baselines.}
    \label{fig:visual_com}
    \vspace{-4mm}
\end{figure*}

\vspace{-1mm}
\noindent \textbf{Qualitative Comparison.}
As shown in Fig.~\ref{fig:visual_com}, LucidNFT recovers more accurate text strokes and finer local textures than its backbones without visibly adding LR-inconsistent content; more examples are given in Appendix~\ref{sec:supp_qualitative}.

\vspace{-6mm}
\subsection{Ablation Study}
\vspace{-3mm}
Table~\ref{tab:ablation} is a controlled ablation on RealLQ250 with LucidFlux, using the same 600-step budget and evaluation as Table~\ref{tab:quantitative}.
By default, LR inputs come from LSDIR~\cite{LSDIR} images degraded by Real-ESRGAN~\cite{Real-ESRGAN} and rewards are fused scalar-first; (A) and (B) are DiffusionNFT baselines with off-the-shelf rewards, and each later row changes a single variable.
\textit{LucidConsistency.} (B)$\to$(C) raises the LucidConsistency gain from $+0.0061$ to $+0.0104$ at a small perceptual cost; since this metric also serves as the reward, the external evidence that it tracks faithfulness better than the frozen embedding is the human alignment in Table~\ref{tab:human_faithfulness}.
\textit{Decoupled normalization.} (C)$\to$(E) lifts the UniPercept gain from $+0.284$ to $+1.773$ and the NIQE reduction from $0.194$ to $0.369$ while consistency also rises, suggesting that the trade-off in (A)--(C) stems largely from scalar-first fusion rather than from the rewards themselves; the training curves of (A), (C), and (E) show the same ordering throughout optimization (Appendix~\ref{sec:supp_curves}).
\textit{LucidLR.} (C)$\to$(D) changes only the training pool and improves all four metrics, indicating that LucidLR is beneficial even without decoupled normalization; combining both in (F) yields the best perceptual results.

\begin{table}[t]
\centering
\captionsetup{font=scriptsize}
\caption{Scalar-first (S) versus decoupled (D) fusion under five reward setups (300 steps, RealLQ250). Entries are gains over LucidFlux; the larger gain of each pair is in bold. ``Wins'' counts the six metrics (Appendix~\ref{sec:supp_reward_grid}) on which decoupled fusion gains more.}
\label{tab:grid_main}
\vspace{1.5mm}
\setlength{\tabcolsep}{3.2pt}
\renewcommand{\arraystretch}{1.1}
\begin{adjustbox}{max width=\linewidth}
\begin{tabular}{l cc cc cc cc cc c}
\toprule
\multirow{2}{*}{\textbf{Reward setup}} & \multicolumn{2}{c}{\textbf{UniPercept IQA}\,$\uparrow$} & \multicolumn{2}{c}{\textbf{CLIP-IQA}\,$\uparrow$} & \multicolumn{2}{c}{\textbf{MUSIQ}\,$\uparrow$} & \multicolumn{2}{c}{\textbf{MANIQA}\,$\uparrow$} & \multicolumn{2}{c}{\textbf{LucidConsistency}\,$\uparrow$} & \multirow{2}{*}{\textbf{Wins}} \\
\cmidrule(lr){2-3}\cmidrule(lr){4-5}\cmidrule(lr){6-7}\cmidrule(lr){8-9}\cmidrule(lr){10-11}
 & S & D & S & D & S & D & S & D & S & D & \\
\midrule
UP-IQA : LC $=1{:}1$ & $+1.21$ & $\mathbf{+1.69}$ & $+0.016$ & $\mathbf{+0.028}$ & $+0.02$ & $\mathbf{+0.55}$ & $-0.002$ & $\mathbf{+0.011}$ & $+0.0144$ & $\mathbf{+0.0170}$ & 6/6 \\
UP-IQA : LC $=2{:}1$ & $+1.01$ & $\mathbf{+1.70}$ & $+0.009$ & $\mathbf{+0.027}$ & $-0.03$ & $\mathbf{+0.69}$ & $-0.006$ & $\mathbf{+0.013}$ & $\mathbf{+0.0183}$ & $+0.0157$ & 5/6 \\
UP-IQA : LC $=1{:}2$ & $+0.78$ & $\mathbf{+0.92}$ & $+0.009$ & $\mathbf{+0.018}$ & $-0.34$ & $\mathbf{+0.10}$ & $-0.010$ & $\mathbf{+0.002}$ & $\mathbf{+0.0175}$ & $+0.0169$ & 5/6 \\
$+$ CLIP-IQA ($1{:}1{:}1$) & $+0.46$ & $\mathbf{+1.31}$ & $+0.004$ & $\mathbf{+0.027}$ & $-0.43$ & $\mathbf{+0.39}$ & $-0.013$ & $\mathbf{+0.011}$ & $+0.0163$ & $\mathbf{+0.0176}$ & 6/6 \\
$+$ Qwen3-VL-Sim ($1{:}1{:}1$) & $+0.91$ & $\mathbf{+0.95}$ & $+0.010$ & $\mathbf{+0.016}$ & $-0.48$ & $\mathbf{+0.15}$ & $-0.010$ & $\mathbf{+0.005}$ & $+0.0163$ & $\mathbf{+0.0172}$ & 6/6 \\
\bottomrule
\end{tabular}
\end{adjustbox}
\vspace{-4mm}
\end{table}

\noindent \textbf{Robustness to reward weights and reward sets.}
Table~\ref{tab:grid_main} compares both fusion schemes under five reward setups with a 300-step budget: three weightings of UniPercept IQA and LucidConsistency, and two extensions with an additional perceptual (CLIP-IQA) or semantic (Qwen3-VL-Sim) reward. Decoupled normalization yields the larger gain on at least five of six metrics in every setup; scalar-first fusion is ahead only on LucidConsistency under unbalanced weights, where some of its perceptual gains (MUSIQ, MANIQA) turn negative (full results in Appendix~\ref{sec:supp_reward_grid}).

\vspace{-5mm}
\section{Conclusion}
\label{sec:con}
\vspace{-4mm}
We presented LucidNFT, which aligns flow-based Real-ISR models toward both perceptual realism and LR faithfulness through an LR-referenced evaluator, decoupled reward normalization, and a large collection of real degraded inputs; it improves perceptual quality on two backbones while a reward-independent human study judges its outputs more LR-faithful.

\vspace{-4mm}
\paragraph{Discussion.}
Our results suggest that hallucination in Real-ISR is capability-dependent: conservative models rarely alter content, whereas stronger models trade LR faithfulness for realism, so hallucination grows more critical as priors improve.

\vspace{-5mm}
\paragraph{Limitations.}
(i)~No reward can compensate for missing evidence under extreme degradations (severe underexposure, heavy blur, or missing regions).
(ii)~Both rewards are learned models that prolonged RL may exploit; the bounded reward weight, KL regularization, and a fixed budget mitigate but do not eliminate reward hacking.
(iii)~LucidLR is biased toward web photography and under-represents domains such as medical, satellite, or industrial imagery.
(iv)~LucidConsistency depends on the frozen Qwen3-VL-Embedding-8B backbone; another backbone requires retraining its LoRA adapters, without a guarantee of comparable performance.
(v)~LucidConsistency is both reward and metric, and no standard reference-free LR-faithfulness benchmark exists; our model-level human study only partially provides external evidence.


\begin{ack}
This work is supported by Guangdong Basic and Applied Basic Research Foundation
(2025A1515140021), and Program of Guangdong Education Department, and
Guangdong Provincial Key Lab of Integrated Communication, Sensing and Computation for Ubiquitous Internet of Things(No.2023B1212010007).
\end{ack}

\bibliographystyle{splncs04}
\bibliography{main}

@String(AAAI  = {AAAI})

@inproceedings{Real-ESRGAN,
  title={{Real-ESRGAN}: Training real-world blind super-resolution with pure synthetic data},
  author={Wang, Xintao and Xie, Liangbin and Dong, Chao and Shan, Ying},
  booktitle={Proceedings of the IEEE/CVF international conference on computer vision},
  pages={1905--1914},
  year={2021}
}

@inproceedings{DIV2K_and_Flickr2K,
  title={{NTIRE} 2017 challenge on single image super-resolution: Dataset and study},
  author={Agustsson, Eirikur and Timofte, Radu},
  booktitle={Proceedings of the IEEE Conference on Computer Vision and Pattern Recognition Workshops},
  pages={126--135},
  year={2017}
}

@inproceedings{RealSR,
  title={Toward real-world single image super-resolution: A new benchmark and a new model},
  author={Cai, Jianrui and Zeng, Hui and Yong, Hongwei and Cao, Zisheng and Zhang, Lei},
  booktitle={Proceedings of the IEEE/CVF International Conference on Computer Vision},
  pages={3086--3095},
  year={2019}
}

@inproceedings{DRealSR,
  title={Component divide-and-conquer for real-world image super-resolution},
  author={Wei, Pengxu and Xie, Ziwei and Lu, Hannan and Zhan, Zongyuan and Ye, Qixiang and Zuo, Wangmeng and Lin, Liang},
  booktitle={Proceedings of the European Conference on Computer Vision},
  pages={101--117},
  year={2020},
  organization={Springer}
}

@article{Resshift,
  title={{ResShift}: Efficient diffusion model for image super-resolution by residual shifting},
  author={Yue, Zongsheng and Wang, Jianyi and Loy, Chen Change},
  journal={Advances in Neural Information Processing Systems},
  volume={36},
  pages={13294--13307},
  year={2023}
}

@inproceedings{SinSR,
  title={{SinSR}: diffusion-based image super-resolution in a single step},
  author={Wang, Yufei and Yang, Wenhan and Chen, Xinyuan and Wang, Yaohui and Guo, Lanqing and Chau, Lap-Pui and Liu, Ziwei and Qiao, Yu and Kot, Alex C and Wen, Bihan},
  booktitle={Proceedings of the IEEE/CVF Conference on Computer Vision and Pattern Recognition},
  pages={25796--25805},
  year={2024}
}

@article{StableSR,
  title={Exploiting diffusion prior for real-world image super-resolution},
  author={Wang, Jianyi and Yue, Zongsheng and Zhou, Shangchen and Chan, Kelvin CK and Loy, Chen Change},
  journal={International Journal of Computer Vision},
  volume={132},
  number={12},
  pages={5929--5949},
  year={2024},
  publisher={Springer}
}

@inproceedings{SeeSR,
  title={{SeeSR}: Towards semantics-aware real-world image super-resolution},
  author={Wu, Rongyuan and Yang, Tao and Sun, Lingchen and Zhang, Zhengqiang and Li, Shuai and Zhang, Lei},
  booktitle={Proceedings of the IEEE/CVF Conference on Computer Vision and Pattern Recognition},
  pages={25456--25467},
  year={2024}
}

@article{DreamClear,
    title={{DreamClear}: High-Capacity Real-World Image Restoration with Privacy-Safe Dataset Curation},
    author={Ai, Yuang and Zhou, Xiaoqiang and Huang, Huaibo and Han, Xiaotian and Chen, Zhengyu and You, Quanzeng and Yang, Hongxia},
    journal={Advances in Neural Information Processing Systems},
    volume={37},
    pages={55443--55469},
    year={2024}
}

@inproceedings{SUPIR,
  title={Scaling Up to Excellence: Practicing Model Scaling for Photo-Realistic Image Restoration In the Wild},
  author={Fanghua Yu and Jinjin Gu and Zheyuan Li and Jinfan Hu and Xiangtao Kong and Xintao Wang and Jingwen He and Yu Qiao and Chao Dong},
  booktitle={Proceedings of the IEEE/CVF Conference on Computer Vision and Pattern Recognition},
  pages={25669--25680},
  year={2024}
}

@inproceedings{CLIP-IQA,
    author = {Wang, Jianyi and Chan, Kelvin CK and Loy, Chen Change},
    title = {Exploring {CLIP} for Assessing the Look and Feel of Images},
    booktitle = {AAAI},
    year = {2023}
}

@inproceedings{Q-Align,
  title={{Q-Align}: Teaching {LMMs} for Visual Scoring via Discrete Text-Defined Levels},
  author={Wu, Haoning and Zhang, Zicheng and Zhang, Weixia and Chen, Chaofeng and Li, Chunyi and Liao, Liang and Wang, Annan and Zhang, Erli and Sun, Wenxiu and Yan, Qiong and Min, Xiongkuo and Zhai, Guangtai and Lin, Weisi},
  booktitle={International Conference on Machine Learning},
  year={2024}
}

@inproceedings{VisualQuality-R1,
  title={{VisualQuality-R1}: Reasoning-Induced Image Quality Assessment via Reinforcement Learning to Rank},
  author={Wu, Tianhe and Zou, Jian and Liang, Jie and Zhang, Lei and Ma, Kede},
  booktitle={Advances in Neural Information Processing Systems},
  year={2025}
}

@article{PSNR,
  title={Image quality assessment: from error visibility to structural similarity},
  author={Wang, Zhou and Bovik, Alan C and Sheikh, Hamid R and Simoncelli, Eero P},
  journal={IEEE Transactions on Image Processing},
  volume={13},
  number={4},
  pages={600--612},
  year={2004},
  publisher={IEEE}
}

@inproceedings{MUSIQ,
  title={{MUSIQ}: Multi-scale image quality transformer},
  author={Ke, Junjie and Wang, Qifei and Wang, Yilin and Milanfar, Peyman and Yang, Feng},
  booktitle={Proceedings of the IEEE/CVF International Conference on Computer Vision},
  pages={5148--5157},
  year={2021}
}

@inproceedings{MANIQA,
  title={{MANIQA}: Multi-dimension Attention Network for No-Reference Image Quality Assessment},
  author={Yang, Sidi and Wu, Tianhe and Shi, Shuwei and Lao, Shanshan and Gong, Yuan and Cao, Mingdeng and Wang, Jiahao and Yang, Yujiu},
  booktitle={Proceedings of the IEEE/CVF Conference on Computer Vision and Pattern Recognition},
  pages={1191--1200},
  year={2022}
}

@ARTICLE{NIQE,
  author={Zhang, Lin and Zhang, Lei and Bovik, Alan C.},
  journal={IEEE Transactions on Image Processing}, 
  title={A Feature-Enriched Completely Blind Image Quality Evaluator}, 
  year={2015},
  volume={24},
  number={8},
  pages={2579-2591}
}

@article{NIMA,
  title={{NIMA}: Neural image assessment},
  author={Talebi, Hossein and Milanfar, Peyman},
  journal={IEEE Transactions on Image Processing},
  volume={27},
  number={8},
  pages={3998--4011},
  year={2018},
  publisher={IEEE}
}

@article{Qwen3-VL-Embedding,
  title={{Qwen3-VL-Embedding} and {Qwen3-VL-Reranker}: A Unified Framework for State-of-the-Art Multimodal Retrieval and Ranking},
  author={Li, Mingxin and Zhang, Yanzhao and Long, Dingkun and Chen, Keqin and Song, Sibo and Bai, Shuai and Yang, Zhibo and Xie, Pengjun and Yang, An and Liu, Dayiheng and others},
  journal={arXiv preprint arXiv:2601.04720},
  year={2026}
}

@inproceedings{dong2014learning,
  title={Learning a deep convolutional network for image super-resolution},
  author={Dong, Chao and Loy, Chen Change and He, Kaiming and Tang, Xiaoou},
  booktitle={Proceedings of the European Conference on Computer Vision},
  pages={184--199},
  year={2014},
  organization={Springer}
}

@inproceedings{zhang2018image,
  title={Image super-resolution using very deep residual channel attention networks},
  author={Zhang, Yulun and Li, Kunpeng and Li, Kai and Wang, Lichen and Zhong, Bineng and Fu, Yun},
  booktitle={Proceedings of the European Conference on Computer Vision},
  pages={286--301},
  year={2018}
}

@inproceedings{chen2023activating,
  title={Activating more pixels in image super-resolution transformer},
  author={Chen, Xiangyu and Wang, Xintao and Zhou, Jiantao and Qiao, Yu and Dong, Chao},
  booktitle={Proceedings of the IEEE/CVF Conference on Computer Vision and Pattern Recognition},
  pages={22367--22377},
  year={2023}
}

@inproceedings{DiffBIR,
  title={{DiffBIR}: Toward blind image restoration with generative diffusion prior},
  author={Lin, Xinqi and He, Jingwen and Chen, Ziyan and Lyu, Zhaoyang and Dai, Bo and Yu, Fanghua and Qiao, Yu and Ouyang, Wanli and Dong, Chao},
  booktitle={European Conference on Computer Vision},
  pages={430--448},
  year={2024},
  organization={Springer}
}

@inproceedings{LucidFlux,
  title={{LucidFlux}: Caption-Free Photo-Realistic Image Restoration via a Large-Scale Diffusion Transformer},
  author={Fei, Song and Ye, Tian and Wang, Lujia and Zhu, Lei},
  booktitle={International Conference on Learning Representations},
  year={2026}
}

@inproceedings{LSDIR,
  title={{LSDIR}: A large scale dataset for image restoration},
  author={Li, Yawei and Zhang, Kai and Liang, Jingyun and Cao, Jiezhang and Liu, Ce and Gong, Rui and Zhang, Yulun and Tang, Hao and Liu, Yun and Demandolx, Denis and Ranjan, Rakesh and Timofte, Radu and Van Gool, Luc},
  booktitle={Proceedings of the IEEE/CVF Conference on Computer Vision and Pattern Recognition},
  pages={1775--1787},
  year={2023}
}

@article{InfoNCE,
  title={Representation learning with contrastive predictive coding},
  author={Oord, Aaron van den and Li, Yazhe and Vinyals, Oriol},
  journal={arXiv preprint arXiv:1807.03748},
  year={2018}
}

@inproceedings{SupCon,
  title={Supervised contrastive learning},
  author={Khosla, Prannay and Teterwak, Piotr and Wang, Chen and Sarna, Aaron and Tian, Yonglong and Isola, Phillip and Maschinot, Aaron and Liu, Ce and Krishnan, Dilip},
  booktitle={Advances in Neural Information Processing Systems},
  volume={33},
  pages={18661--18673},
  year={2020}
}

@inproceedings{SD3,
  title={Scaling rectified flow transformers for high-resolution image synthesis},
  author={Esser, Patrick and Kulal, Sumith and Blattmann, Andreas and Entezari, Rahim and M{\"u}ller, Jonas and Saini, Harry and Levi, Yam and Lorenz, Dominik and Sauer, Axel and Boesel, Frederic and others},
  booktitle={Forty-first International Conference on Machine Learning},
  year={2024}
}

@inproceedings{SD,
  title={High-resolution image synthesis with latent diffusion models},
  author={Rombach, Robin and Blattmann, Andreas and Lorenz, Dominik and Esser, Patrick and Ommer, Bj{\"o}rn},
  booktitle={Proceedings of the IEEE/CVF Conference on Computer Vision and Pattern Recognition},
  pages={10684--10695},
  year={2022}
}

@misc{FLUX,
    author={{Black Forest Labs}},
    title={{FLUX}},
    year={2024}
}

@inproceedings{SDXL,
  title={{SDXL}: Improving latent diffusion models for high-resolution image synthesis},
  author={Podell, Dustin and English, Zion and Lacey, Kyle and Blattmann, Andreas and Dockhorn, Tim and M{\"u}ller, Jonas and Penna, Joe and Rombach, Robin},
  booktitle={International Conference on Learning Representations},
  year={2024}
}

@inproceedings{PixArt-alpha,
  title={{PixArt-$\alpha$}: Fast Training of Diffusion Transformer for Photorealistic Text-to-Image Synthesis},
  author={Junsong Chen and Jincheng Yu and Chongjian Ge and Lewei Yao and Enze Xie and Yue Wu and Zhongdao Wang and James Kwok and Ping Luo and Huchuan Lu and Zhenguo Li},
  booktitle={International Conference on Learning Representations},
  year={2024}
}

@inproceedings{DiT4SR,
  title={{DiT4SR}: Taming Diffusion Transformer for Real-World Image Super-Resolution},
  author={Duan, Zheng-Peng and Zhang, Jiawei and Jin, Xin and Zhang, Ziheng and Xiong, Zheng and Zou, Dongqing and Ren, Jimmy and Guo, Chun-Le and Li, Chongyi},
  booktitle={Proceedings of the IEEE/CVF International Conference on Computer Vision},
  year={2025}
}

@inproceedings{UniPercept,
  title={{UniPercept}: Towards Unified Perceptual-Level Image Understanding across Aesthetics, Quality, Structure, and Texture},
  author={Shuo Cao and Jiayang Li and Xiaohui Li and Yuandong Pu and Kaiwen Zhu and Yuanting Gao and Siqi Luo and Yi Xin and Qi Qin and Yu Zhou and Xiangyu Chen and Wenlong Zhang and Bin Fu and Yu Qiao and Yihao Liu},
  booktitle={International Conference on Machine Learning},
  year={2026}
}

@inproceedings{LoRA,
  title={{LoRA}: Low-rank adaptation of large language models},
  author={Hu, Edward J. and Shen, Yelong and Wallis, Phillip and Allen-Zhu, Zeyuan and Li, Yuanzhi and Wang, Shean and Wang, Lu and Chen, Weizhu},
  booktitle={International Conference on Learning Representations},
  year={2022}
}

@inproceedings{AdamW,
  title={Decoupled weight decay regularization},
  author={Loshchilov, Ilya and Hutter, Frank},
  booktitle={International Conference on Learning Representations},
  year={2019}
}

@inproceedings{DiffusionNFT,
  title={{DiffusionNFT}: Online diffusion reinforcement with forward process},
  author={Zheng, Kaiwen and Chen, Huayu and Ye, Haotian and Wang, Haoxiang and Zhang, Qinsheng and Jiang, Kai and Su, Hang and Ermon, Stefano and Zhu, Jun and Liu, Ming-Yu},
  booktitle={International Conference on Learning Representations},
  year={2026}
}

@book{RL,
  title={Reinforcement learning: An introduction},
  author={Sutton, Richard S and Barto, Andrew G and others},
  volume={1},
  year={1998},
  publisher={MIT Press}
}

@article{cohen2024looks,
  title={Looks too good to be true: An information-theoretic analysis of hallucinations in generative restoration models},
  author={Cohen, Regev and Kligvasser, Idan and Rivlin, Ehud and Freedman, Daniel},
  journal={Advances in Neural Information Processing Systems},
  volume={37},
  pages={22596--22623},
  year={2024}
}

@article{DPO,
  title={Direct preference optimization: Your language model is secretly a reward model},
  author={Rafailov, Rafael and Sharma, Archit and Mitchell, Eric and Manning, Christopher D and Ermon, Stefano and Finn, Chelsea},
  journal={Advances in neural information processing systems},
  volume={36},
  pages={53728--53741},
  year={2023}
}

@inproceedings{DiffusionDPO,
  title={Diffusion model alignment using direct preference optimization},
  author={Wallace, Bram and Dang, Meihua and Rafailov, Rafael and Zhou, Linqi and Lou, Aaron and Purushwalkam, Senthil and Ermon, Stefano and Xiong, Caiming and Joty, Shafiq and Naik, Nikhil},
  booktitle={Proceedings of the IEEE/CVF Conference on Computer Vision and Pattern Recognition},
  pages={8228--8238},
  year={2024}
}

@inproceedings{Flow-GRPO,
  title={{Flow-GRPO}: Training flow matching models via online {RL}},
  author={Liu, Jie and Liu, Gongye and Liang, Jiajun and Li, Yangguang and Liu, Jiaheng and Wang, Xintao and Wan, Pengfei and Zhang, Di and Ouyang, Wanli},
  booktitle={Advances in Neural Information Processing Systems},
  year={2025}
}

@article{DanceGRPO,
  title={{DanceGRPO}: Unleashing {GRPO} on visual generation},
  author={Xue, Zeyue and Wu, Jie and Gao, Yu and Kong, Fangyuan and Zhu, Lingting and Chen, Mengzhao and Liu, Zhiheng and Liu, Wei and Guo, Qiushan and Huang, Weilin and others},
  journal={arXiv preprint arXiv:2505.07818},
  year={2025}
}

@inproceedings{lipman2022flow,
  title={Flow matching for generative modeling},
  author={Lipman, Yaron and Chen, Ricky TQ and Ben-Hamu, Heli and Nickel, Maximilian and Le, Matt},
  booktitle={International Conference on Learning Representations},
  year={2023}
}

@article{liu2022flow,
  title={Flow straight and fast: Learning to generate and transfer data with rectified flow},
  author={Liu, Xingchao and Gong, Chengyue and Liu, Qiang},
  journal={arXiv preprint arXiv:2209.03003},
  year={2022}
}

@inproceedings{DDPO,
  title={Training diffusion models with reinforcement learning},
  author={Black, Kevin and Janner, Michael and Du, Yilun and Kostrikov, Ilya and Levine, Sergey},
  booktitle={International Conference on Learning Representations},
  year={2024}
}

@article{PPO,
  title={Proximal policy optimization algorithms},
  author={Schulman, John and Wolski, Filip and Dhariwal, Prafulla and Radford, Alec and Klimov, Oleg},
  journal={arXiv preprint arXiv:1707.06347},
  year={2017}
}

@inproceedings{SAGI,
  title={{SAGI}: Semantically Aligned and Uncertainty Guided {AI} Image Inpainting},
  author={Giakoumoglou, Paschalis and Karageorgiou, Dimitrios and Papadopoulos, Symeon and Petrantonakis, Panagiotis C.},
  booktitle={Proceedings of the IEEE/CVF International Conference on Computer Vision},
  year={2025}
}

@inproceedings{RAISE,
author = {Dang-Nguyen, Duc-Tien and Pasquini, Cecilia and Conotter, Valentina and Boato, Giulia},
title = {{RAISE}: A raw images dataset for digital image forensics},
year = {2015},
publisher = {Association for Computing Machinery},
address = {New York, NY, USA},
doi = {10.1145/2713168.2713194},
booktitle = {Proceedings of the 6th ACM Multimedia Systems Conference},
pages = {219--224},
series = {MMSys '15}
}

@inproceedings{DP2O-SR,
  title={{DP$^2$O-SR}: Direct Perceptual Preference Optimization for Real-World Image Super-Resolution},
  author={Wu, Rongyuan and Sun, Lingchen and Zhang, Zhengqiang and Wang, Shihao and Wu, Tianhe and Yi, Qiaosi and Li, Shuai and Zhang, Lei},
  booktitle={Advances in Neural Information Processing Systems},
  year={2025}
}

@article{RefReward-SR,
  title={{RefReward-SR}: {LR-Conditioned} Reward Modeling for Preference-Aligned Super-Resolution},
  author={Song, Yushuai and Quan, Weize and Wang, Weining and Sun, Jiahui and Liu, Jing and Li, Meng and Yu, Pengbin and Chen, Zhentao and Shen, Wei and Yuan, Lunxi and others},
  journal={arXiv preprint arXiv:2603.24198},
  year={2026}
}

@inproceedings{FinPercep-RM,
  title={{FinPercep-RM}: A Fine-grained Reward Model and Co-evolutionary Curriculum for {RL-based} Real-world Super-Resolution},
  author={Liu, Yidi and Fan, Zihao and Huang, Jie and Xiao, Jie and Li, Dong and Zhang, Wenlong and Bai, Lei and Fu, Xueyang and Zha, Zheng-Jun},
  booktitle={Proceedings of the IEEE/CVF Conference on Computer Vision and Pattern Recognition},
  pages={4839--4849},
  year={2026}
}

@inproceedings{RealSR-R1,
  title={{RealSR-R1}: Reinforcement Learning for Real-World Image Super-Resolution with Vision-Language Chain-of-Thought},
  author={Qiao, Junbo and Cai, Miaomiao and Li, Wei and Liu, Yutong and Huang, Xudong and He, Gaoqi and Xie, Jiao and Hu, Jie and Chen, Xinghao and Lin, Shaohui},
  booktitle={arXiv preprint arXiv:2506.16796},
  year={2025}
}

@inproceedings{XPSR,
  title={{XPSR}: Cross-modal Priors for Diffusion-based Image Super-Resolution},
  author={Qu, Yunpeng and Yuan, Kun and Zhao, Kai and Xie, Qizhi and Hao, Jinhua and Sun, Ming and Zhou, Chao},
  booktitle={European Conference on Computer Vision},
  pages={285--303},
  year={2024}
}

@article{LPNSR,
  title={{LPNSR}: Prior-Enhanced Diffusion Image Super-Resolution via {LR-Guided} Noise Prediction},
  author={Huang, Shuwei and Liu, Shizhuo and Wei, Zijun},
  journal={arXiv preprint arXiv:2603.21045},
  year={2026}
}

@article{DSPO,
  title={{DSPO}: Direct semantic preference optimization for real-world image super-resolution},
  author={Cai, Miaomiao and Li, Simiao and Li, Wei and Huang, Xudong and Chen, Hanting and Hu, Jie and Wang, Yunhe},
  journal={arXiv preprint arXiv:2504.15176},
  year={2025}
}

@article{GDPO-SR,
  title={{GDPO-SR}: Group Direct Preference Optimization for One-Step Generative Image Super-Resolution},
  author={Yi, Qiaosi and Li, Shuai and Wu, Rongyuan and Sun, Lingchen and Zhang, Zhengqiang and Zhang, Lei},
  journal={arXiv preprint arXiv:2603.16769},
  year={2026}
}

@article{OARS,
  title={{OARS}: Process-Aware Online Alignment for Generative Real-World Image Super-Resolution},
  author={Zhao, Shijie and Zhang, Xuanyu and Chen, Bin and Li, Weiqi and Xing, Qunliang and Zhang, Kexin and Wang, Yan and Li, Junlin and Zhang, Li and Zhang, Jian and others},
  journal={arXiv preprint arXiv:2603.12811},
  year={2026}
}

@article{Lens,
  title={Lens: Rethinking Training Efficiency for Foundational Text-to-Image Models},
  author={Chen, Dong and Wei, Fangyun and Wan, Ziyu and Chen, Dongdong and Zhang, Jiawei and Zhao, Jinjing and Zhang, Sirui and Yue, Yang and Liang, Zhiyang and Guo, Baining and Luo, Chong and Bao, Jianmin and Li, Ji and Shi, Lei and Yang, Qinhong and Wu, Xiuyu and Feng, Xuelu and Lu, Yan and Dong, Yanchen and Wang, Yitong and Chen, Yunuo},
  journal={arXiv preprint arXiv:2605.21573},
  year={2026}
}

@article{Seedance,
  title={Seedance 1.0: Exploring the Boundaries of Video Generation Models},
  author={Gao, Yu and Guo, Haoyuan and Hoang, Tuyen and Huang, Weilin and Jiang, Lu and others},
  journal={arXiv preprint arXiv:2506.09113},
  year={2025}
}

@inproceedings{InstructGPT,
  title={Training Language Models to Follow Instructions with Human Feedback},
  author={Ouyang, Long and Wu, Jeffrey and Jiang, Xu and Almeida, Diogo and Wainwright, Carroll and Mishkin, Pamela and Zhang, Chong and Agarwal, Sandhini and Slama, Katarina and Ray, Alex and others},
  booktitle={Advances in Neural Information Processing Systems},
  year={2022}
}

@article{UniWorld-V2,
  title={{UniWorld-V2}: Reinforce Image Editing with Diffusion Negative-aware Finetuning and {MLLM} Implicit Feedback},
  author={Li, Zongjian and Liu, Zheyuan and Zhang, Qihui and Lin, Bin and Wu, Feize and Yuan, Shenghai and Yan, Zhiyuan and Ye, Yang and Yu, Wangbo and Niu, Yuwei and Wang, Shaodong and Cheng, Xinhua and Yuan, Li},
  journal={arXiv preprint arXiv:2510.16888},
  year={2025}
}

@article{FireRed-Image-Edit,
  title={{FireRed-Image-Edit-1.0} Technical Report},
  author={{Super Intelligence Team}},
  journal={arXiv preprint arXiv:2602.13344},
  year={2026}
}

@article{Flow-Factory,
  title={{Flow-Factory}: A Unified Framework for Reinforcement Learning in Flow-Matching Models},
  author={Ping, Bowen and Jia, Chengyou and Luo, Minnan and Qian, Hangwei and Tsang, Ivor},
  journal={arXiv preprint arXiv:2602.12529},
  year={2026}
}

@article{MeanFlowNFT,
  title={{MeanFlowNFT}: Bringing Forward-Process {RL} to Average-Velocity Generators},
  author={Huang, Yushi and Zhou, Xiangxin and Zhang, Jun and Bo, Liefeng and Pang, Tianyu},
  journal={arXiv preprint arXiv:2607.15273},
  year={2026}
}

@inproceedings{BrushNet,
  title={{BrushNet}: A Plug-and-Play Image Inpainting Model with Decomposed Dual-Branch Diffusion},
  author={Ju, Xuan and Liu, Xian and Wang, Xintao and Bian, Yuxuan and Shan, Ying and Xu, Qiang},
  booktitle={European Conference on Computer Vision},
  year={2024}
}

@inproceedings{ControlNet,
  title={Adding Conditional Control to Text-to-Image Diffusion Models},
  author={Zhang, Lvmin and Rao, Anyi and Agrawala, Maneesh},
  booktitle={Proceedings of the IEEE/CVF International Conference on Computer Vision},
  pages={3836--3847},
  year={2023}
}

@inproceedings{HD-Painter,
  title={{HD-Painter}: High-Resolution and Prompt-Faithful Text-Guided Image Inpainting with Diffusion Models},
  author={Manukyan, Hayk and Sargsyan, Andranik and Atanyan, Barsegh and Wang, Zhangyang and Navasardyan, Shant and Shi, Humphrey},
  booktitle={International Conference on Learning Representations},
  year={2025}
}

@inproceedings{PowerPaint,
  title={A Task is Worth One Word: Learning with Task Prompts for High-Quality Versatile Image Inpainting},
  author={Zhuang, Junhao and Zeng, Yanhong and Liu, Wenran and Yuan, Chun and Chen, Kai},
  booktitle={European Conference on Computer Vision},
  year={2024}
}

@article{InpaintAnything,
  title={Inpaint Anything: Segment Anything Meets Image Inpainting},
  author={Yu, Tao and Feng, Runseng and Feng, Ruoyu and Liu, Jinming and Jin, Xin and Zeng, Wenjun and Chen, Zhibo},
  journal={arXiv preprint arXiv:2304.06790},
  year={2023}
}

@article{LandisKoch,
  title={The Measurement of Observer Agreement for Categorical Data},
  author={Landis, J. Richard and Koch, Gary G.},
  journal={Biometrics},
  volume={33},
  number={1},
  pages={159--174},
  year={1977}
}

@inproceedings{SnowMaster,
  title={{SnowMaster}: Comprehensive Real-world Image Desnowing via {MLLM} with Multi-Model Feedback Optimization},
  author={Lai, Jianyu and Chen, Sixiang and Lin, Yunlong and Ye, Tian and Liu, Yun and Fei, Song and Xing, Zhaohu and Wu, Hongtao and Wang, Weiming and Zhu, Lei},
  booktitle={Proceedings of the IEEE/CVF Conference on Computer Vision and Pattern Recognition},
  pages={4302--4312},
  year={2025}
}

@inproceedings{PosterCraft,
  title={{PosterCraft}: Rethinking High-Quality Aesthetic Poster Generation in a Unified Framework},
  author={Chen, Sixiang and Lai, Jianyu and Gao, Jialin and Ye, Tian and Chen, Haoyu and Shi, Hengyu and Shao, Shitong and Lin, Yunlong and Fei, Song and Xing, Zhaohu and Jin, Yeying and Luo, Junfeng and Wei, Xiaoming and Zhu, Lei},
  booktitle={International Conference on Learning Representations},
  year={2026}
}

@inproceedings{PosterOmni,
  title={{PosterOmni}: Generalized Artistic Poster Creation via Task Distillation and Unified Reward Feedback},
  author={Chen, Sixiang and Lai, Jianyu and Gao, Jialin and Shi, Hengyu and Liu, Zhongying and Ye, Tian and Luo, Junfeng and Wei, Xiaoming and Zhu, Lei},
  booktitle={Proceedings of the IEEE/CVF Conference on Computer Vision and Pattern Recognition},
  pages={5978--5987},
  year={2026}
}

@inproceedings{PosterReward,
  title={{PosterReward}: Unlocking Accurate Evaluation for High-Quality Graphic Design Generation},
  author={Lai, Jianyu and Chen, Sixiang and Gao, Jialin and Shi, Hengyu and Liu, Zhongying and Zhai, Fuxiang and Luo, Junfeng and Wei, Xiaoming and Wang, Lujia and Zhu, Lei},
  booktitle={Proceedings of the IEEE/CVF Conference on Computer Vision and Pattern Recognition},
  pages={7762--7772},
  year={2026}
}

@inproceedings{UltraFlux,
  title={{UltraFlux}: Data-Model Co-Design for High-quality Native {4K} Text-to-Image Generation across Diverse Aspect Ratios},
  author={Ye, Tian and Fei, Song and Zhu, Lei},
  booktitle={Proceedings of the IEEE/CVF Conference on Computer Vision and Pattern Recognition},
  pages={22070--22079},
  year={2026}
}

@inproceedings{DTPM,
  title={Learning Diffusion Texture Priors for Image Restoration},
  author={Ye, Tian and Chen, Sixiang and Chai, Wenhao and Xing, Zhaohu and Qin, Jing and Lin, Ge and Zhu, Lei},
  booktitle={Proceedings of the IEEE/CVF Conference on Computer Vision and Pattern Recognition},
  pages={2524--2534},
  year={2024}
}

@inproceedings{GenHaze,
  title={{GenHaze}: Pioneering Controllable One-Step Realistic Haze Generation for Real-World Dehazing},
  author={Chen, Sixiang and Ye, Tian and Lin, Yunlong and Jin, Yeying and Yang, Yijun and Chen, Haoyu and Lai, Jianyu and Fei, Song and Xing, Zhaohu and Tsung, Fugee and Zhu, Lei},
  booktitle={Proceedings of the IEEE/CVF International Conference on Computer Vision},
  pages={9194--9205},
  year={2025}
}

@inproceedings{S2DLight,
  title={{S2DLight}: Single Step Diffusion Network via Degradation-Aware Adaptation for Surgical Endoscopic Image Low-Light Enhancement},
  author={Fei, Song and Wang, Guanyi and Ye, Tian and Chen, Sixiang and Lai, Jianyu and Zhu, Lei},
  booktitle={International Conference on Medical Image Computing and Computer-Assisted Intervention},
  pages={444--454},
  year={2026}
}

@inproceedings{MoSE,
  title={{MoSE}: Mixture-of-Scale-Experts for Medical Image Restoration},
  author={Li, Xingyu and Xing, Zhaohu and Liu, Zekai and Li, Shuaibo and Wang, Hongqiu and Fei, Song and Zhu, Lei},
  booktitle={International Conference on Medical Image Computing and Computer-Assisted Intervention},
  pages={306--316},
  year={2026}
}

@inproceedings{RestoreAgent,
  title={{RestoreAgent}: Autonomous Image Restoration Agent via Multimodal Large Language Models},
  author={Chen, Haoyu and Li, Wenbo and Gu, Jinjin and Ren, Jingjing and Chen, Sixiang and Ye, Tian and Pei, Renjing and Zhou, Kaiwen and Song, Fenglong and Zhu, Lei},
  booktitle={Advances in Neural Information Processing Systems},
  volume={37},
  pages={110643--110666},
  year={2024}
}

@inproceedings{COCO,
  title={Microsoft {COCO}: Common Objects in Context},
  author={Lin, Tsung-Yi and Maire, Michael and Belongie, Serge and Hays, James and Perona, Pietro and Ramanan, Deva and Doll{\'a}r, Piotr and Zitnick, C. Lawrence},
  booktitle={European Conference on Computer Vision},
  pages={740--755},
  year={2014}
}

@article{Fleiss,
  title={Measuring nominal scale agreement among many raters},
  author={Fleiss, Joseph L.},
  journal={Psychological Bulletin},
  volume={76},
  number={5},
  pages={378--382},
  year={1971}
}

@inproceedings{DPMSolver,
  title={{DPM-Solver}: A Fast {ODE} Solver for Diffusion Probabilistic Model Sampling in Around 10 Steps},
  author={Lu, Cheng and Zhou, Yuhao and Bao, Fan and Chen, Jianfei and Li, Chongxuan and Zhu, Jun},
  booktitle={Advances in Neural Information Processing Systems},
  volume={35},
  pages={5775--5787},
  year={2022}
}

@article{DeepSeekV4,
  title={{DeepSeek-V4}: Towards Highly Efficient Million-Token Context Intelligence},
  author={{DeepSeek-AI}},
  journal={arXiv preprint arXiv:2606.19348},
  year={2026}
}


\appendix

\clearpage

\section{Additional Related Work}
\label{sec:supp_related}

\paragraph{Preference and reward-based Real-ISR alignment.}
Recent works have explored preference or reward optimization specifically for Real-ISR. 
DSPO~\cite{DSPO} applies semantic preference optimization, DP$^2$O-SR~\cite{DP2O-SR} studies direct perceptual preference optimization, GDPO-SR~\cite{GDPO-SR} introduces group DPO for one-step generative ISR, and OARS~\cite{OARS} performs process-aware online alignment with an MLLM-based reward. 
RefReward-SR~\cite{RefReward-SR} and FinPercep-RM~\cite{FinPercep-RM} further study LR-conditioned or fine-grained reward modeling.

DP$^2$O-SR is the most closely related preference-based Real-ISR work to our setting. 
It constructs hybrid IQA rewards, selects winner--loser pairs from stochastic SR candidates, and optimizes a DPO-style hierarchical preference objective. 
LucidNFT targets a different bottleneck in flow-matching Real-ISR: LR-referenced faithfulness is explicitly modeled by LucidConsistency, and perceptual and faithfulness rewards are normalized separately within each LR-conditioned rollout group before fusion. 
This preserves objective-wise reward contrast, rather than collapsing heterogeneous objectives into pairwise scalar preferences.

A direct comparison with DP$^2$O-SR is not feasible: its public release provides inference scripts and checkpoints for C-SD2/C-FLUX, but not the training code, training data, or IQA reward labels required to reproduce its post-training pipeline. 
In addition, its released flow-based artifact follows a ControlNet-FLUX-style pipeline, whereas our controlled experiments use LucidFlux as the base model. 
A direct DP$^2$O-SR-on-LucidFlux comparison would therefore require a substantial re-implementation of rollout generation, IQA labeling, pair selection, and HPO-weighted training. 
We instead report a same-backbone Diffusion-DPO~\cite{DiffusionDPO} implementation, a form of preference optimization also used in recent generation pipelines~\cite{PosterCraft}, to isolate direct preference optimization under our data, backbone, and training setting.

\paragraph{Learned evaluators and reward feedback.}
Beyond Real-ISR, learned evaluators and reward feedback have been used to steer generative and restoration models toward task-specific criteria.
SnowMaster~\cite{SnowMaster} uses an MLLM evaluator refined by multi-model feedback optimization to assess real-world desnowing results and filter pseudo-labels, PosterReward~\cite{PosterReward} models the typography and layout quality of graphic designs, and PosterOmni~\cite{PosterOmni} trains a unified reward model to guide reinforcement learning for poster creation.
LucidConsistency plays the corresponding role for LR faithfulness in Real-ISR, where the reference is a degraded input rather than a text prompt.

\paragraph{LR-guided generative super-resolution.}
Several non-RL approaches improve LR anchoring through model design. 
XPSR~\cite{XPSR} introduces cross-modal priors for diffusion-based SR, LPNSR~\cite{LPNSR} strengthens LR-guided noise prediction to reduce hallucination, and related restoration models learn dedicated diffusion texture priors~\cite{DTPM}. 
These methods modify the restoration prior or conditioning mechanism, whereas LucidNFT aligns stochastic SR rollouts through reward-guided fine-tuning; as it leaves the backbone architecture unchanged, it could also be applied to stronger flow priors, such as FLUX-based models scaled to native 4K generation~\cite{UltraFlux}.

\paragraph{Real-world image super-resolution datasets.}
Training and evaluating Real-ISR models requires data that reflect realistic degradation patterns. 
Synthetic pipelines such as Real-ESRGAN~\cite{Real-ESRGAN} generate blur, noise, and compression artifacts from high-quality datasets such as DIV2K, Flickr2K~\cite{DIV2K_and_Flickr2K}, and LSDIR~\cite{LSDIR}, and generative models can synthesize more realistic degradations such as haze~\cite{GenHaze}. 
Real-captured paired datasets such as RealSR~\cite{RealSR} and DRealSR~\cite{DRealSR} are valuable benchmarks, but their scale and capture diversity are limited. 
LucidLR is designed for a different role: it provides large-scale unpaired real LR inputs for RL fine-tuning, where diverse degradations are needed to induce informative rollout variation and reward contrast.
Specialized domains such as surgical endoscopy~\cite{S2DLight} and medical imaging~\cite{MoSE} involve degradations that web photographs do not cover, which motivates limitation (iii) in Sec.~\ref{sec:con}.

\section{LucidConsistency: Data, Human Alignment, and Robustness}
\label{sec:supp_lc}
\subsection{Training Data Construction}
\label{sec:supp_degradation_settings}

\paragraph{Content-consistent pools.} We generate degradation views using a controlled Real-ESRGAN-style pipeline. 
For each HR crop, we produce three ordered degradation levels: mild, medium, and heavy. 
The clean crop is saved as the GT view, while each degraded LR image is bicubic-upsampled back to the GT resolution to form the corresponding degraded view of the content-consistent pool.

\paragraph{Original--inpainted pairs.} For hard negatives, we use the fully regenerated subset of SAGI~\cite{SAGI}. 
Each sample consists of an original image and its inpainted counterpart. 
SAGI inpainting methods output images at their own working resolution (typically around 1K pixels on the long side), which can differ substantially from that of the original photograph (up to about 5K pixels for RAISE). We therefore first bring each pair to a common pixel grid: the original is resized to the resolution of its inpainted counterpart, so that both images depict the same scene at the same scale, and the original is never compared against an upsampled, artificially smoothed inpainted image. 
If the short side of the pair is below the crop size, both images are bicubically upscaled by the same factor rather than padded. 
Both images then share the same random crop coordinates within each epoch, and the crop is sampled so that it contains the inpainted region given by the SAGI edit mask; the paired crops are therefore identical outside the edit and differ only in the locally regenerated content. 
Their degradation operations are sampled independently, so the paired views remain spatially aligned while covering different blur, noise, resizing, and compression realizations.

\paragraph{Cropping.} All HR crops use size $1024\times1024$. 
For each source image we generate four crops (one per epoch), each at the three degradation levels. 
Different epochs use different crop regions, while paired original--inpainted samples share crop coordinates within the same epoch. Content-consistent pools follow the same rule for small images: images whose short side is below 1024 pixels are bicubically upscaled before cropping instead of being padded.

\paragraph{Degradation levels.}
Each degradation level randomly samples Real-ESRGAN operations from a predefined range:
\begin{itemize}
    \item \textbf{Mild}: weak blur, mostly keeping the original resolution or slight downsampling, very light Gaussian/Poisson noise, high JPEG quality, and no second-order degradation.
    \item \textbf{Medium}: moderate blur, stronger downsampling, moderate noise, medium JPEG compression, and occasional second-order degradation.
    \item \textbf{Heavy}: strong blur, aggressive downsampling, stronger noise, lower JPEG quality, and always applying second-order degradation.
\end{itemize}

Table~\ref{tab:supp_degradation} lists the parameter ranges.
\begin{table}[h]
\centering
\caption{Parameter ranges of the three Real-ESRGAN degradation levels used to build content-consistent pools.}
\label{tab:supp_degradation}
\small
\begin{tabular}{lccc}
\toprule
\textbf{Parameter} & \textbf{Mild} & \textbf{Medium} & \textbf{Heavy} \\
\midrule
Blur sigma & $[0.2,0.45]$ & $[0.6,1.5]$ & $[1.2,3.0]$ \\
Resize range & $[0.9,1.0]$ & $[0.5,0.8]$ & $[0.25,0.6]$ \\
Noise range & $[0,1.5]$ & $[3,8]$ & $[8,20]$ \\
Poisson scale & $[0.0,0.12]$ & $[0.3,1.0]$ & $[1.0,2.5]$ \\
JPEG quality & $[92,100]$ & $[50,85]$ & $[20,60]$ \\
Second-order prob. & $0.0$ & $0.2$ & $1.0$ \\
\bottomrule
\end{tabular}
\end{table}

The pipeline follows the Real-ESRGAN two-stage degradation design with blur, resizing, noise, JPEG compression, and sinc filtering. 
For the medium and heavy levels, the second-stage degradation uses the same noise and JPEG ranges as the first stage, while the second-stage resize ranges are $[0.5,0.8]$ and $[0.25,0.8]$, respectively.

\subsection{Human Alignment Protocol}
\label{sec:supp_human_pref_lr_faithfulness}

We provide additional details for the human-aligned LR-faithfulness evaluation in Sec.~\ref{sec:lc_eval}. 
The study is conducted on RealSR~\cite{RealSR}. 
For each LR image, we collect SR candidates from eight generative SR methods: ResShift~\cite{Resshift}, SeeSR~\cite{SeeSR}, LucidFlux~\cite{LucidFlux}, DreamClear~\cite{DreamClear}, SUPIR~\cite{SUPIR}, DiffBIRv2~\cite{DiffBIR}, DiT4SR~\cite{DiT4SR}, and StableSR~\cite{StableSR}. 
Each annotation task presents the LR image and four randomly sampled SR candidates from this candidate pool.

\begin{figure}[t]
\centering
\includegraphics[width=\linewidth]{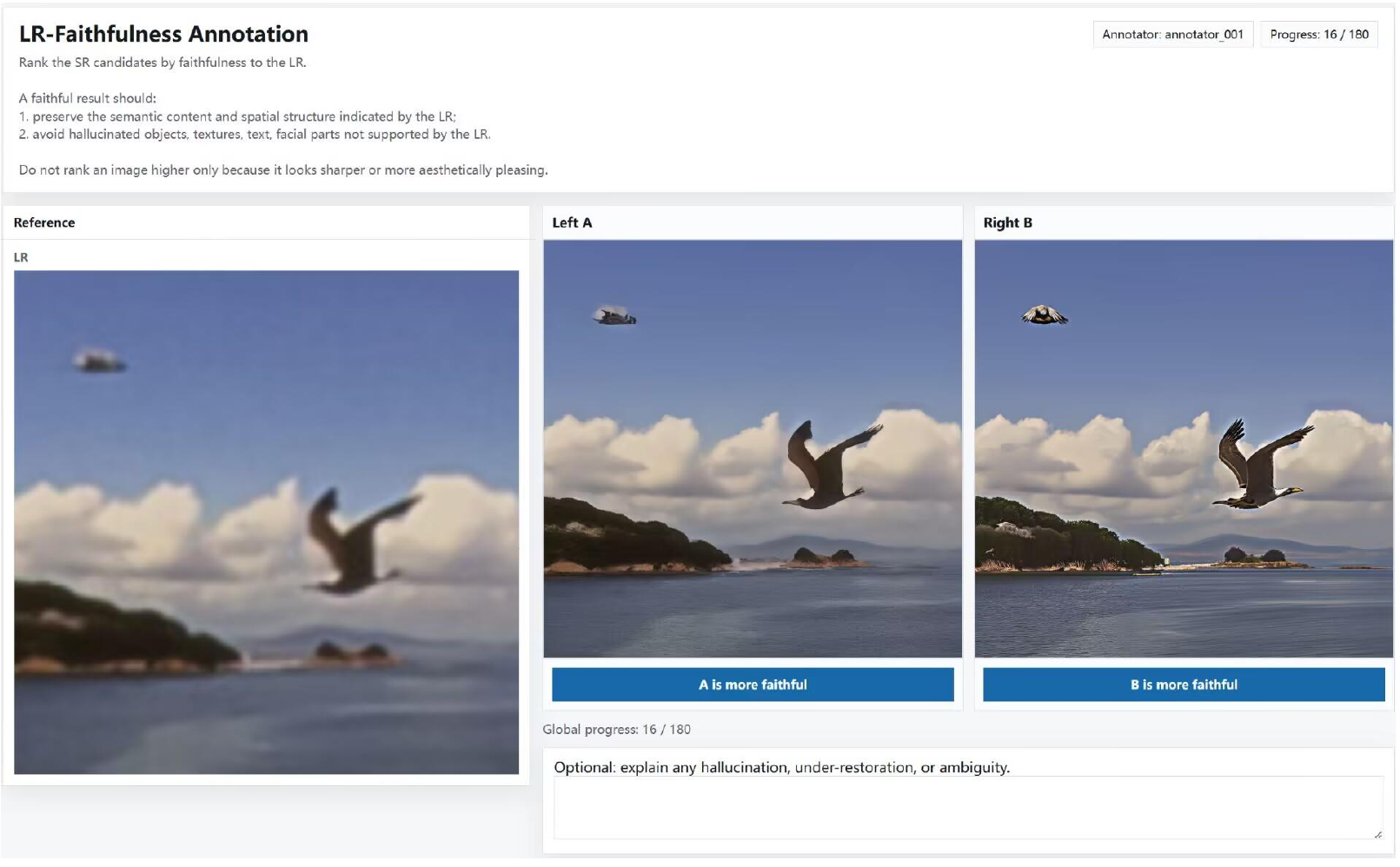}
\caption{
Annotation interface for LR-faithfulness evaluation. 
Annotators are shown the LR image and four randomly ordered SR candidates, and rank the candidates by faithfulness to the LR evidence rather than perceptual sharpness alone.
}
\label{fig:human_lr_interface}
\end{figure}

Five expert annotators are asked to judge LR-faithfulness rather than perceptual sharpness or aesthetics. 
Specifically, annotators are instructed to rank candidates according to whether the restored result preserves LR-supported semantic content and spatial structure, while avoiding hallucinated objects, textures, text, facial parts, or repeated patterns unsupported by the LR observation. 
The annotation interface is shown in Fig.~\ref{fig:human_lr_interface}. 
Each four-candidate ranking is decomposed into six pairwise preferences, and candidate order is randomized to reduce presentation bias.

We evaluate automatic metrics by comparing their induced rankings with the aggregated human preferences. 
We report three metrics: \textbf{Agreement}, the pairwise agreement with human preferences; \textbf{Recall@1}, whether the metric selects the human-preferred best candidate; and \textbf{Filter@1}, whether the metric assigns the lowest score to the human-labeled worst candidate. 
As reported in the main paper, LucidConsistency achieves higher agreement with human LR-faithfulness judgments than perceptual metrics and the frozen Qwen3-VL-Embedding baseline.

\subsection{Leave-One-Inpainter-Out Evaluation}
\label{sec:supp_loio}

\paragraph{Protocol.} Our hard-negative pool contains 11 single-round combinations of four inpainting methods and seven checkpoints from the SD~1.5, SD~2, and SDXL families, plus two-round edits (Sec.~\ref{sec:impl}). The SDXL family is used only by BrushNet, with the SDXL base model and with Juggernaut XL. Removing every pair in which either of these two generators takes part, in either editing round (2,784 of the 37.7K pairs: 1,437 single-round and 1,347 two-round), therefore withholds an entire base architecture, which is the strongest single-fold test of architectural generalization available in our training pool. We retrain LucidConsistency on the remaining pairs under otherwise identical settings (LOIO) and compare it with the evaluator trained on all pairs (Full).

\paragraph{Test sets.} (i) \emph{SAGI test split of the held-out SDXL-family generators}, disjoint from training. For an original image $x^{o}$ with such an inpainting $x^{i}$, a faithful evaluator should score the content-preserving pair $(x^{o},D(x^{o}))$ above the content-altered pair $(x^{o},x^{i})$, where $D(\cdot)$ is a Real-ESRGAN synthetic degradation at a mild, medium, or heavy level. (ii) \emph{Human-annotated set}: the 40 RealSR LR inputs of Appendix~\ref{sec:supp_model_human}, each paired with the two real SR outputs, with the human majority vote as ground truth. On both sets we report paired agreement, i.e., the fraction of pairs on which the evaluator's preference matches the ground truth.

\begin{table}[h]
\centering
\caption{Paired agreement of LucidConsistency with and without the SDXL-family generators in training.}
\label{tab:supp_loio}
\small
\begin{tabular}{lcc}
\toprule
\textbf{Evaluator} & \textbf{SAGI test split} $\uparrow$ & \textbf{Human-annotated set} $\uparrow$ \\
\midrule
Full (all generators) & 0.74 & 0.69 \\
LOIO (SDXL family held out) & 0.70 & 0.67 \\
\bottomrule
\end{tabular}
\end{table}

\section{LucidNFT: Implementation and Training Dynamics}
\label{sec:supp_nft}
\subsection{Hyperparameters and Pseudocode}
\label{sec:supp_impl}

Table~\ref{tab:supp_hparams} lists the full LucidNFT training configuration used for both backbones in the main experiments, and Algorithm~\ref{alg:lucidnft} summarizes one training iteration. Relative to DiffusionNFT~\cite{DiffusionNFT}, only the mapping from the reward vector to the reward weight (step 3) differs; the policy objective is unchanged.

\begin{table}[h]
\centering
\caption{LucidNFT training hyperparameters.}
\label{tab:supp_hparams}
\small
\begin{tabular}{ll}
\toprule
\textbf{Item} & \textbf{Setting} \\
\midrule
Hardware / RL steps & 8 NVIDIA A100 GPUs / 600 steps \\
Rollout resolution & $768\times768$ \\
LR groups & 3 per GPU (24 globally) \\
Rollouts per LR input ($M$) & 12 \\
Rewards and weights & UniPercept IQA ($\lambda_p=1.0$), LucidConsistency ($\lambda_f=1.0$) \\
LoRA rank / alpha & 32 / 64 \\
Optimizer & AdamW, lr $3\times10^{-5}$, $\beta_1=0.9$, $\beta_2=0.999$, weight decay $10^{-4}$ \\
NFT interpolation weight $\beta_{\mathrm{NFT}}$ & 1.0 \\
KL regularization weight $\lambda_{\mathrm{KL}}$ & $10^{-4}$ \\
Old-policy mixing $\eta_t$ (Alg.~\ref{alg:lucidnft}) & $\min(10^{-3}t,\,0.5)$ \\
EMA of policy weights (evaluation) & decay 0.9 \\
$Z_c$ (advantage clip) / grad-norm clip & 5.0 / 1.0 \\
Timestep-training fraction & 0.99 \\
Update epochs per rollout batch & 1 (no rollout replay) \\
Sampler & DPM-Solver-2~\cite{DPMSolver}, 6 steps (training) / 15 steps (validation) \\
Rollout noise level & 0.7 (deterministic after injection) \\
Guidance scale / precision & 4 / bf16 (matched to LucidFlux) \\
\bottomrule
\end{tabular}
\end{table}

\begin{algobox}[label={alg:lucidnft}]{One LucidNFT Training Iteration}
\textbf{Input:} policy $v_\theta$, sampling policy $v_{\text{old}}$ (no gradient), rewards $\{R^{(k)}\}_{k=1}^{K}$ with weights $\{\lambda_k\}$, LR batch $\{x_i\}$.
\begin{enumerate}[leftmargin=*,nosep]
    \item \textbf{Rollout.} For each LR input $x_i$, sample $M$ restorations $\{y_{i,j}\}_{j=1}^{M}$ from $v_{\text{old}}$ with the black-box solver.
    \item \textbf{Score.} Compute the reward vector $\mathbf{r}_{i,j}=(R^{(1)}(x_i,y_{i,j}),\ldots,R^{(K)}(x_i,y_{i,j}))$.
    \item \textbf{Decoupled normalization.} Normalize each reward within its rollout group (Eq.~\eqref{eq:dim_norm}), fuse $\tilde a_{i,j}=\sum_k\lambda_k z^{(k)}_{i,j}$, standardize over the batch, clip, and map to $r^{L}_{i,j}\in[0,1]$ (Eq.~\eqref{eq:reward_map}).
    \item \textbf{NFT update.} Sample $t$ and noise, form $x_t$ from $y_{i,j}$, and minimize Eq.~\eqref{eq:diffnft_obj} with $r=r^{L}_{i,j}$, plus the KL term weighted by $\lambda_{\mathrm{KL}}$; one epoch over the fresh rollouts.
    \item \textbf{Old-policy update.} Set $v_{\text{old}}\leftarrow\eta_t v_{\text{old}}+(1-\eta_t)v_\theta$ with $\eta_t=\min(10^{-3}t,0.5)$.
\end{enumerate}
\end{algobox}

\subsection{Advantage Separability}
\label{sec:supp_adv}
\begin{figure}[h]
    \centering
    \includegraphics[width=0.78\linewidth]{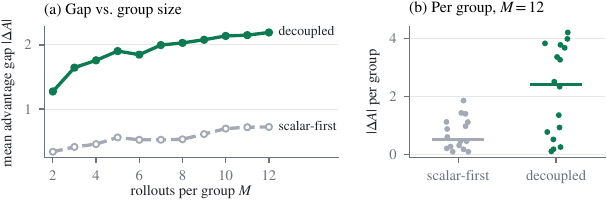}
    \caption{Advantage gap $|\Delta A|$ between the two candidates with the largest reward difference in a group (16 LR inputs, 12 LucidFlux rollouts each, same data as Fig.~\ref{fig:mechanism}). (a)~Mean over groups versus the number of rollouts $M$. (b)~Individual groups at $M=12$; bars mark the median.}
    \label{fig:advantage_ana}
\end{figure}

Fig.~\ref{fig:mechanism}c counts how many distinct advantage levels a group contains. Fig.~\ref{fig:advantage_ana} complements it with the size of the contrast: for each group of the same 16 LR inputs, we take the two candidates whose rewards differ most and measure the gap between their advantages. Scalar-first fusion keeps this gap small, 0.3--0.7, because candidates that win on one reward and lose on the other obtain similar sums; decoupled normalization makes it three to four times larger for every group size (1.27 vs.\ 0.34 at $M=2$, 2.19 vs.\ 0.72 at $M=12$) and larger in 12 of the 16 individual groups, so the most different candidates receive markedly different update weights.

\subsection{Training Dynamics}
\label{sec:supp_curves}
\begin{figure*}[t]
    \centering
    \includegraphics[width=0.92\linewidth]{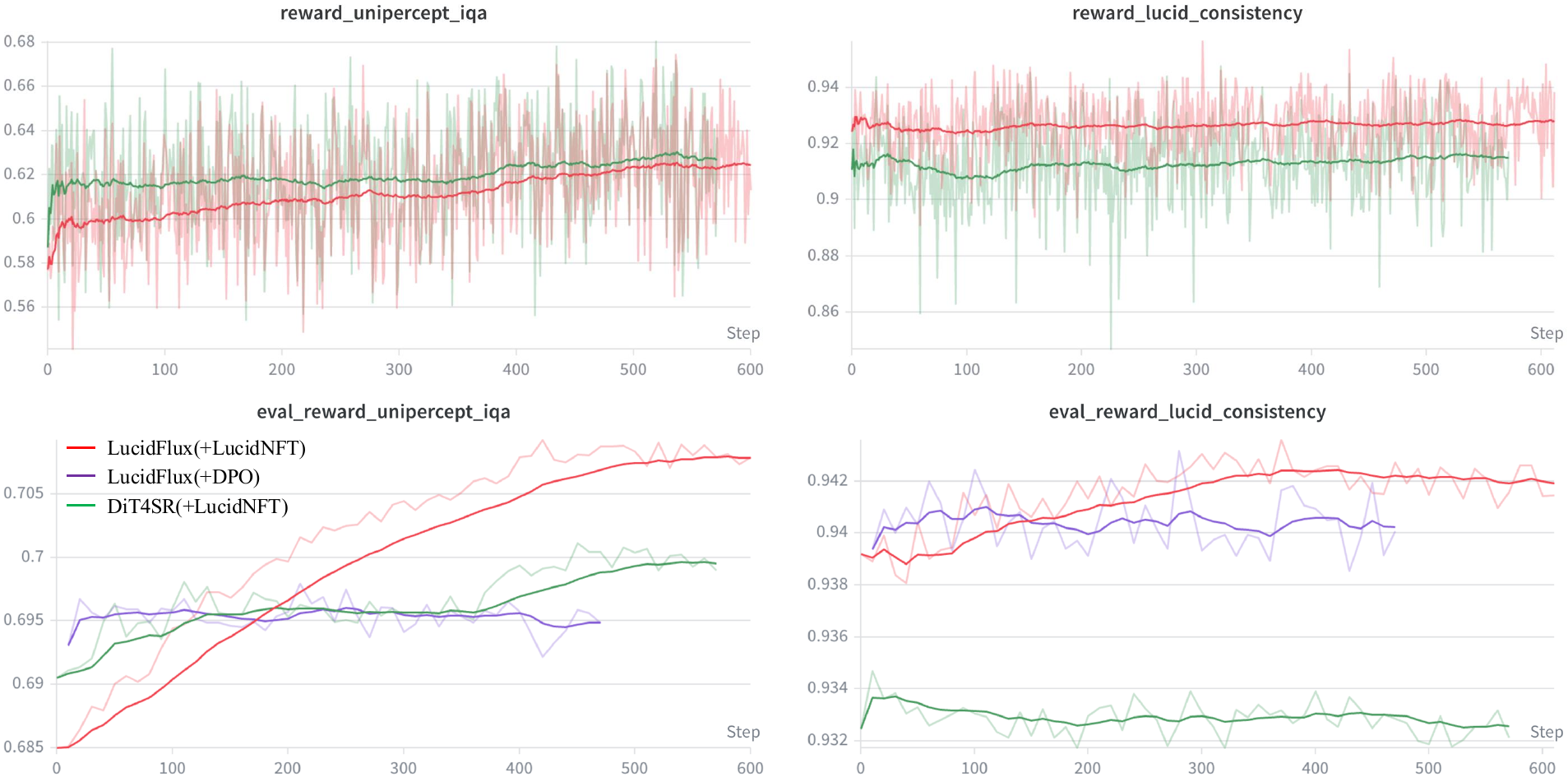}
    \captionsetup{font=scriptsize}
    \caption{Training curves of LucidNFT and DPO fine-tuning.
    Curves report fast validation rewards computed with reduced inference steps. LucidNFT steadily improves perceptual rewards while keeping LucidConsistency stable; the final full evaluation in Table~\ref{tab:quantitative} shows improved LR-faithfulness over the corresponding backbones.}
    \label{fig:curve_lucidflux}
\end{figure*}

\begin{figure*}[t]
    \centering
    \includegraphics[width=\linewidth]{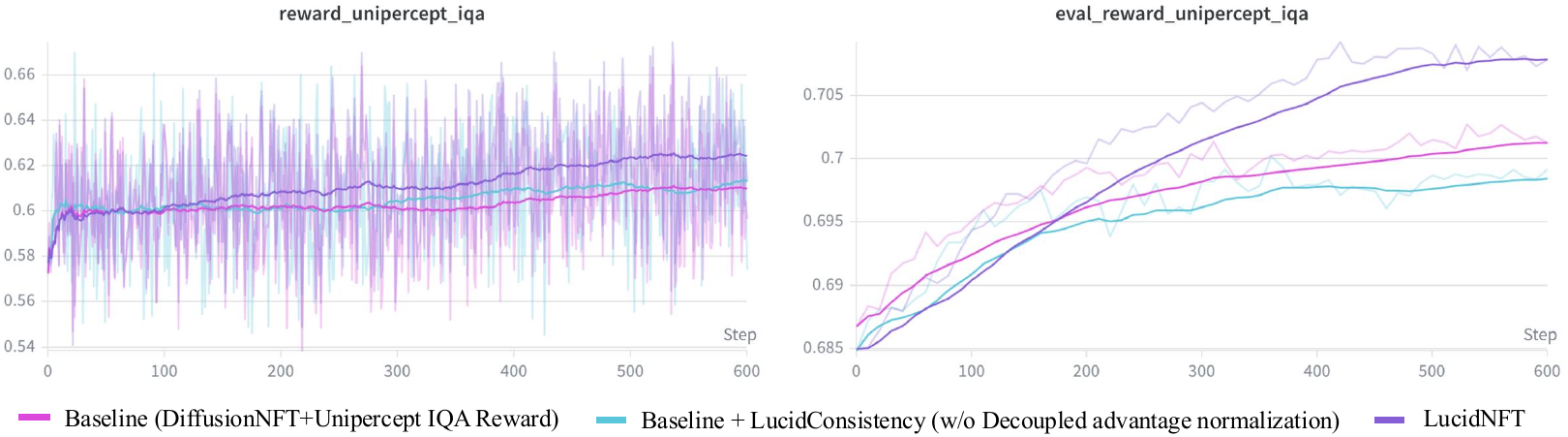}
    \caption{UniPercept IQA reward during training for three LucidFlux variants of Table~\ref{tab:ablation}: IQA-only RL (A), scalar-first fusion with LucidConsistency (C), and decoupled normalization (E).}
    \label{fig:curve_lucidflux_supp}
\end{figure*}

Fig.~\ref{fig:curve_lucidflux} shows the training dynamics of LucidNFT and Diffusion-DPO. Validation rewards are computed with fewer sampling steps than the final evaluation, and LucidConsistency varies less than the perceptual reward during training; the final evaluation in Table~\ref{tab:quantitative} shows that LucidNFT improves LucidConsistency over the corresponding backbones.
Fig.~\ref{fig:curve_lucidflux_supp} compares three reward formulations from Table~\ref{tab:ablation}. IQA-only RL (A) raises the perceptual reward above the pretrained baseline; adding LucidConsistency through scalar-first fusion (C) lowers it again, reflecting the perceptual--faithfulness trade-off; and decoupled normalization (E) yields the highest perceptual reward of the three, consistent with the controlled ablation.

\section{Additional Results}
\label{sec:supp_results}
\subsection{Model-Level Human Preference Study}
\label{sec:supp_model_human}

The study in Sec.~\ref{sec:effectiveness} compares trained models directly, without any LucidConsistency scores. We use 40 held-out RealSR~\cite{RealSR} LR inputs. For each input, the outputs of LucidFlux and LucidFlux(+LucidNFT) are shown side by side with the LR image, in random order. Five expert annotators are asked to select the output that is better supported by the LR content, explicitly disregarding sharpness, color saturation, and aesthetic quality; ties are not allowed. Every annotator judges all 40 pairs, giving 200 binary judgments.

\begin{table}[h]
\centering
\caption{Model-level human preference on LR faithfulness (RealSR).}
\label{tab:supp_model_human}
\small
\begin{tabular}{lccccc}
\toprule
\textbf{Benchmark} & \textbf{Annotators} & \textbf{Judgments} & \textbf{LucidFlux(+LucidNFT)} & \textbf{LucidFlux} & \textbf{Fleiss' $\kappa$} \\
\midrule
RealSR & 5 & 200 & \textbf{72.5\%} & 27.5\% & 0.415 \\
\bottomrule
\end{tabular}
\end{table}

A Fleiss' $\kappa$~\cite{Fleiss} of 0.415 corresponds to moderate agreement on the Landis--Koch scale~\cite{LandisKoch}, indicating that the preference reflects consensus across independent annotators rather than a narrow-majority effect.

\subsection{Relative Improvements over the LucidFlux Backbone}
\label{sec:supp_relative}

Table~\ref{tab:supp_relative} reports the relative change of LucidFlux(+DPO) and LucidFlux(+LucidNFT) over the LucidFlux backbone for every metric and benchmark of Table~\ref{tab:quantitative}. For NIQE, a negative change denotes improvement.

\begin{table}[h]
\centering
\caption{Relative change (\%) over the same LucidFlux backbone. DPO denotes Diffusion-DPO; UP-IQA denotes UniPercept IQA; LC denotes LucidConsistency.}
\label{tab:supp_relative}
\small
\setlength{\tabcolsep}{3pt}
\begin{adjustbox}{max width=\linewidth}
\begin{tabular}{llccccccccc}
\toprule
\textbf{Benchmark} & \textbf{Method} & \textbf{CLIP-IQA+}$\uparrow$ & \textbf{Q-Align}$\uparrow$ & \textbf{MUSIQ}$\uparrow$ & \textbf{MANIQA}$\uparrow$ & \textbf{CLIP-IQA}$\uparrow$ & \textbf{NIQE}$\downarrow$ & \textbf{UP-IQA}$\uparrow$ & \textbf{VQ-R1}$\uparrow$ & \textbf{LC}$\uparrow$ \\
\midrule
\multirow{2}{*}{RealLQ250} & +DPO & +0.28 & +0.86 & +0.16 & +0.59 & +0.90 & +1.00 & +0.29 & +0.37 & +0.64 \\
 & +LucidNFT & +3.57 & +1.82 & +1.54 & +4.13 & +5.51 & $-$13.04 & +3.59 & +2.28 & +1.17 \\
\midrule
\multirow{2}{*}{DRealSR} & +DPO & +0.21 & +0.72 & $-$0.07 & $-$0.19 & +0.60 & $-$1.18 & $-$0.21 & $-$0.32 & +0.87 \\
 & +LucidNFT & +5.39 & +3.45 & +5.50 & +6.97 & +5.99 & $-$17.65 & +6.47 & +3.75 & +0.75 \\
\midrule
\multirow{2}{*}{RealSR} & +DPO & +0.39 & +1.08 & +0.06 & +0.37 & +1.07 & $-$2.76 & +0.64 & +0.31 & +0.89 \\
 & +LucidNFT & +7.23 & +3.07 & +3.93 & +8.08 & +9.07 & $-$17.88 & +7.77 & +4.87 & +0.79 \\
\midrule
\multirow{2}{*}{Average} & +DPO & +0.29 & +0.89 & +0.05 & +0.26 & +0.86 & $-$0.98 & +0.24 & +0.12 & +0.80 \\
 & +LucidNFT & +5.39 & +2.78 & +3.66 & +6.39 & +6.87 & $-$16.19 & +5.94 & +3.63 & +0.90 \\
\bottomrule
\end{tabular}
\end{adjustbox}
\end{table}

\subsection{Robustness to Reward Weights and Reward Sets}
\label{sec:supp_reward_grid}

We train ten configurations on LucidLR for 300 RL steps with all other hyperparameters as in Table~\ref{tab:supp_hparams}, and evaluate on RealLQ250: five reward setups, each with scalar-first fusion and with decoupled normalization. The setups are three weightings $\lambda_p{:}\lambda_f\in\{1{:}1,2{:}1,1{:}2\}$ of UniPercept IQA (UP-IQA) and LucidConsistency (LC), and two three-reward extensions that add a perceptual reward (CLIP-IQA) or a semantic reward (Qwen3-VL-Sim, the cosine similarity between LR and SR under the frozen Qwen3-VL-Embedding-8B backbone). All ten runs share the same budget, isolating the relative behavior of the two fusion schemes. ``Decoupled wins'' counts, per configuration, the metrics on which decoupled normalization yields the larger gain; it summarizes directional consistency across correlated metrics rather than independent statistical tests.

\begin{table}[h]
\centering
\caption{Reward-configuration grid on RealLQ250. The first row is absolute; other rows report changes from the backbone (larger is better on all columns).}
\label{tab:supp_reward_grid}
\small
\setlength{\tabcolsep}{3pt}
\begin{adjustbox}{max width=\linewidth}
\begin{tabular}{llccccccc}
\toprule
\textbf{Configuration} & \textbf{Fusion} & \textbf{UP-IQA}$\uparrow$ & \textbf{CLIP-IQA}$\uparrow$ & \textbf{CLIP-IQA+}$\uparrow$ & \textbf{MUSIQ}$\uparrow$ & \textbf{MANIQA}$\uparrow$ & \textbf{LC}$\uparrow$ & \textbf{Decoupled wins} \\
\midrule
LucidFlux backbone & -- & 70.93 & 0.686 & 0.721 & 72.34 & 0.523 & 0.9237 & -- \\
\midrule
\multirow{2}{*}{UP-IQA : LC = 1:1} & scalar-first & +1.21 & +0.016 & +0.005 & +0.02 & $-$0.002 & +0.0144 & \multirow{2}{*}{6/6} \\
 & decoupled & \textbf{+1.69} & \textbf{+0.028} & \textbf{+0.015} & \textbf{+0.55} & \textbf{+0.011} & \textbf{+0.0170} & \\
\midrule
\multirow{2}{*}{UP-IQA : LC = 2:1} & scalar-first & +1.01 & +0.009 & +0.002 & $-$0.03 & $-$0.006 & \textbf{+0.0183} & \multirow{2}{*}{5/6} \\
 & decoupled & \textbf{+1.70} & \textbf{+0.027} & \textbf{+0.016} & \textbf{+0.69} & \textbf{+0.013} & +0.0157 & \\
\midrule
\multirow{2}{*}{UP-IQA : LC = 1:2} & scalar-first & +0.78 & +0.009 & $-$0.002 & $-$0.34 & $-$0.010 & \textbf{+0.0175} & \multirow{2}{*}{5/6} \\
 & decoupled & \textbf{+0.92} & \textbf{+0.018} & \textbf{+0.005} & \textbf{+0.10} & \textbf{+0.002} & +0.0169 & \\
\midrule
\multirow{2}{*}{UP-IQA : LC : CLIP-IQA = 1:1:1} & scalar-first & +0.46 & +0.004 & $-$0.005 & $-$0.43 & $-$0.013 & +0.0163 & \multirow{2}{*}{6/6} \\
 & decoupled & \textbf{+1.31} & \textbf{+0.027} & \textbf{+0.011} & \textbf{+0.39} & \textbf{+0.011} & \textbf{+0.0176} & \\
\midrule
\multirow{2}{*}{UP-IQA : LC : Qwen3-VL-Sim = 1:1:1} & scalar-first & +0.91 & +0.010 & $-$0.004 & $-$0.48 & $-$0.010 & +0.0163 & \multirow{2}{*}{6/6} \\
 & decoupled & \textbf{+0.95} & \textbf{+0.016} & \textbf{+0.004} & \textbf{+0.15} & \textbf{+0.005} & \textbf{+0.0172} & \\
\bottomrule
\end{tabular}
\end{adjustbox}
\end{table}

\subsection{Additional Qualitative Results}
\label{sec:supp_qualitative}
\begin{figure*}[t]
    \centering
    \includegraphics[page=2, width=\linewidth]{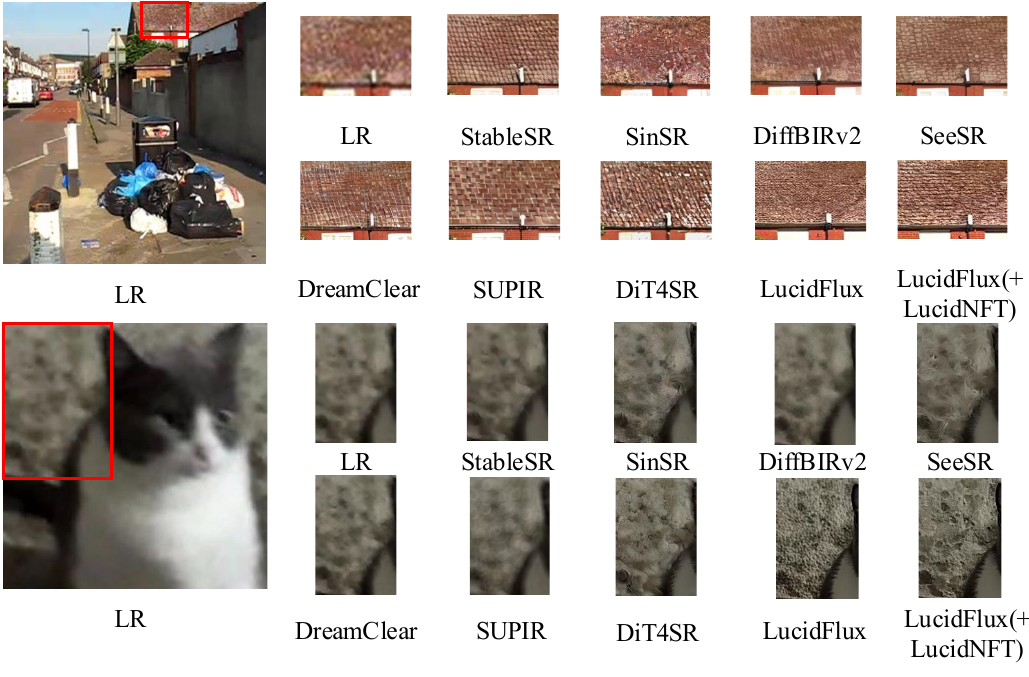}
    \caption{Additional qualitative comparisons on RealLQ250~\cite{DreamClear}. Compared with the LucidFlux baseline, LucidFlux(+LucidNFT) recovers clearer local details while better preserving LR-supported structures.}
    \label{fig:visual_com_supp}
\end{figure*}

Fig.~\ref{fig:visual_com_supp} shows further RealLQ250 examples of LucidFlux and LucidFlux(+LucidNFT). The fine-tuned model restores sharper local structures, such as text strokes and regular patterns, while keeping them consistent with the LR input; these examples are illustrative, and the quantitative and human evaluations in Sec.~\ref{sec:effectiveness} remain the primary evidence.

\section{LucidLR Dataset}
\label{sec:supp_lucidlr}
\subsection{Collection, Screening, and Release}
\label{sec:supp_lucidlr_release}

\paragraph{Collection.} Images are retrieved from the Wikimedia Commons categories \emph{Images of low quality} and \emph{Blurred images} through the official API (\url{https://commons.wikimedia.org}). The raw pool contains 21,698 unique files.

\paragraph{Content screening.} Every image is screened by DeepSeek-V4.1-Flash at temperature 0, after downscaling (never upscaling) so that the longer side is at most 1024 pixels. A first pass labels each image as safe or as one of \emph{nudity/sexual}, \emph{suggestive}, \emph{gore/violence}, \emph{drugs}, \emph{hate symbols}, or \emph{other}. Images flagged by the first pass are re-examined by a narrower second pass that separates photographic content from depictions in art (e.g., classical nudity or religious and battle paintings) and corrects false alarms such as dark scenes. Table~\ref{tab:supp_lucidlr_filter} lists the resulting filtering steps; artistic depictions are kept.

\begin{table}[h]
\centering
\caption{Filtering steps from the collected pool to the released LucidLR.}
\label{tab:supp_lucidlr_filter}
\small
\begin{tabular}{lr}
\toprule
\textbf{Step} & \textbf{Images} \\
\midrule
Pool after de-duplication, excluding a 1,000-image held-out split & 20,698 \\
$-$ explicit sexual content, photographic gore, personal data & $-$56 \\
$-$ hate symbols & $-$16 \\
$-$ short side below 128 pixels & $-$719 \\
$-$ source file deleted from Commons (license unverifiable) & $-$169 \\
$+$ screened images collected from the same categories & $+$262 \\
\midrule
\textbf{Released LucidLR} & \textbf{20,000} \\
\bottomrule
\end{tabular}
\end{table}

\paragraph{Licensing.} LucidLR includes only images whose Commons licenses permit redistribution. Of the 20,000 images, 12,542 are under CC~BY-SA (2.0/2.5/3.0/4.0 and ported variants), 2,780 under CC~BY, 4,469 are public domain, CC0, or otherwise free of restrictions, and 209 carry other free licenses (e.g., GFDL or Free Art License). Any image whose redistribution status is uncertain (e.g., trademark, personality-rights, or freedom-of-panorama concerns) is excluded from the image bundle and kept as a URL-only entry in the metadata index, so that users can still reproduce the collection while responsibility for local redistribution remains with them.

\paragraph{Release format.} We release LucidLR in two forms: (1) a pre-fetched image bundle, since the Commons API enforces per-user rate limits that make re-downloading 20K images slow; and (2) a metadata index with a reference downloader, \texttt{download\_lucidlr.py}, for users who prefer to fetch the images themselves. Every image keeps its original license and is accompanied by a JSON sidecar with the following fields: image name; direct URL and source page; uploader, upload time, and original date; width and height; author; license, license URL, and license slug; usage terms; description; and categories. An attribution index generated from the sidecars satisfies the attribution and share-alike terms.

\subsection{Statistics and Comparison with Real-ISR Benchmarks}
\label{sec:supp_lucidlr_stats}
We characterize LucidLR by its semantic content, resolution, and degradations, and compare it with three representative real-world low-quality benchmarks, RealLQ250, RealSR, and DRealSR, annotated with the same model and taxonomy. All statistics refer to the 20,000 released images.

\begin{figure*}[t]
  \centering
  \includegraphics[width=\linewidth]{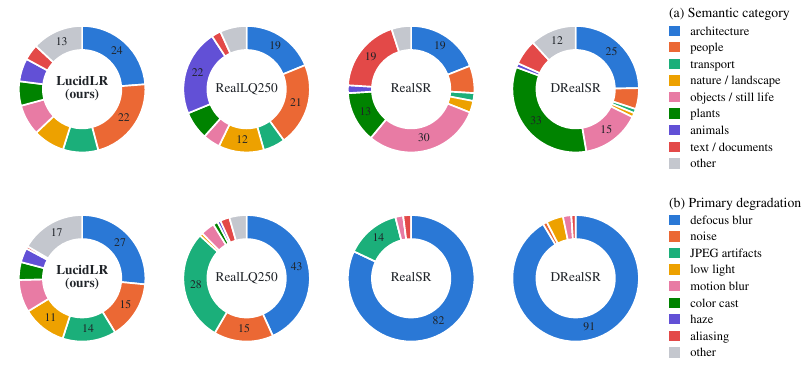}
  \caption{(a)~Semantic category and (b)~primary degradation of LucidLR and three real-world benchmarks, annotated with the same model and taxonomy. Each category has the same color and position in every chart; categories outside the eight shown are merged into \emph{other}, and wedges of at least 10\% give their percentage. The benchmarks are dominated by a few scene types and by defocus blur, whereas LucidLR spreads over many categories and degradations.}
  \label{fig:supp_lucidlr_stats}
\end{figure*}

\paragraph{Semantic content and degradations.} Multimodal large language models have been used to identify degradations and plan restoration~\cite{RestoreAgent}; here, the same DeepSeek-V4.1-Flash call that screens each image also assigns one of 13 semantic categories, a single \emph{primary} degradation, and the set of all visible degradation types (prompt at the end of this section). Fig.~\ref{fig:supp_lucidlr_stats} compares the resulting distributions. LucidLR covers all 13 semantic categories without a dominant one (architecture 23.8\%, people 22.1\%, and no other category above 10\%), whereas the benchmarks concentrate on a few scene types, e.g., objects, text, and architecture in RealSR. The degradation statistics show the same contrast: defocus blur is the primary degradation in 82.0\% and 91.4\% of the RealSR and DRealSR images and in 43.2\% of RealLQ250, but in only 26.5\% of LucidLR, whose primary degradations spread over defocus blur, noise, JPEG artifacts, low light, motion blur, color cast, haze, and overexposure. Counting all visible degradations rather than only the primary one, LucidLR images frequently combine several of them, and types such as overexposure, lens flare, and occlusion, which occur in at most 2.0\% of the images of any benchmark, appear in 3.0--9.8\% of LucidLR. These annotations characterize the datasets and are not used for training; because all images are stored as JPEG, codec damage is almost always labeled \texttt{jpeg\_artifacts}, so this label and \texttt{compression\_artifacts} should be read together.

\paragraph{Resolution.} LucidLR is not resized to a common size: every image keeps the resolution at which it was collected, so resolutions vary widely (Table~\ref{tab:supp_resolution}). The median image is $1080\times960$ (1.05\,MP) with a median short side of 853 pixels; 65.7\% of the images have a short side of at least 512 pixels and 46.0\% of at least 1024 pixels, while the long tail reaches 172\,MP. 70.2\% of the images are landscape, 28.4\% portrait, and 1.3\% square. This spread of resolutions complements the fixed-size crops of existing benchmarks.

\begin{table}[h]
\centering
\caption{Short-side distribution of the 20,000 LucidLR images.}
\label{tab:supp_resolution}
\small
\setlength{\tabcolsep}{4pt}
\begin{tabular}{lcccccccc}
\toprule
\textbf{Short side (px)} & $<$256 & 256--384 & 384--512 & 512--768 & 768--1024 & 1024--1536 & 1536--2048 & $\ge$2048 \\
\midrule
\textbf{Images} & 2,068 & 1,720 & 3,064 & 2,528 & 1,416 & 2,752 & 2,597 & 3,855 \\
\textbf{Share} & 10.3\% & 8.6\% & 15.3\% & 12.6\% & 7.1\% & 13.8\% & 13.0\% & 19.3\% \\
\bottomrule
\end{tabular}
\end{table}

\paragraph{Annotation prompt.}
The first screening pass, the semantic category, and the degradation labels come from a single call with the following prompt, used verbatim for all four datasets.

\begin{promptbox}{Screening and Annotation of LucidLR Images}
You are a dataset curator preparing a public image dataset of real-world degraded photographs. Analyze the single input image and return ONE JSON object.

Definitions of the degradation labels (use these exact strings):
\begin{itemize}[left=0pt,nosep]
    \item \textbf{motion\_blur}: directional blur caused by camera or object motion
    \item \textbf{defocus\_blur}: out-of-focus blur or shallow depth-of-field
    \item \textbf{noise}: visible sensor noise, grain, chroma noise
    \item \textbf{jpeg\_artifacts}: JPEG blocking, mosquito noise, DCT artifacts
    \item \textbf{compression\_artifacts}: generic codec artifacts beyond JPEG
    \item \textbf{aliasing}: jagged edges or resampling artifacts
    \item \textbf{haze / fog}: atmospheric scattering reducing contrast
    \item \textbf{rain\_streaks / snow}: visible precipitation structures
    \item \textbf{low\_light}: severe underexposure or dark scenes
    \item \textbf{overexposure}: clipped highlights or blown-out regions
    \item \textbf{color\_cast}: global color tint due to white balance failure
    \item \textbf{color\_distortion}: abnormal color reproduction
    \item \textbf{dirty\_lens}: stains, droplets, or smears on the lens
    \item \textbf{occlusion}: partial obstruction by objects
    \item \textbf{lens\_flare}: bright streaks or glare caused by strong light
    \item \textbf{other}: degradation exists but does not match the above labels
    \item \textbf{none}: image is mostly clean with no obvious degradation
\end{itemize}

Semantic categories (choose the category covering the most salient subject; if people occupy the main subject area, choose people):
\begin{itemize}[left=0pt,nosep]
    \item \textbf{people}: humans as main subject (portraits, crowds, ceremonies, workers)
    \item \textbf{architecture}: buildings, monuments, bridges, streets, cityscapes
    \item \textbf{nature\_landscape}: mountains, forests, water, sky, fields, beaches
    \item \textbf{animals}: any non-human animal as main subject
    \item \textbf{plants}: flowers, leaves, trees as the principal subject (close-ups)
    \item \textbf{food}: dishes, ingredients, drinks
    \item \textbf{text\_document}: signs, posters, book pages, maps, screenshots, forms where text dominates
    \item \textbf{art\_illustration}: paintings, drawings, engravings, cartoons, logos, graphics, renders
    \item \textbf{objects\_still\_life}: equipment, products, tools, close-ups of inanimate objects
    \item \textbf{indoor\_scene}: interiors, rooms, museums, stage settings
    \item \textbf{transport}: vehicles, trains, aircraft, ships, roads as main subject
    \item \textbf{events\_sports}: sports, performances, concerts, military parades
    \item \textbf{other}: none of the above
\end{itemize}

Return exactly this JSON, no markdown, no commentary:

\begin{lstlisting}[language=json,basicstyle=\ttfamily\scriptsize,columns=fullflexible]
{
  "nsfw": <true if the image contains explicit nudity, sexual content, sex acts,
           graphic gore or graphic violence; suggestive but non-explicit content is NOT nsfw>,
  "nsfw_type": one of [none, nudity_sexual, suggestive, gore_violence, drugs,
                hate_symbols, other],
  "nsfw_confidence": <0.0-1.0>,
  "is_photograph": <true for camera/photographic images; false for drawings, paintings,
                    diagrams, maps, logos, screenshots, 3D renders>,
  "semantic_category": one of the semantic categories above,
  "primary_degradation": <one label from the degradation list, the dominant one>,
  "degradation_types": <1-6 labels from the degradation list, strongest first;
                        if the image is mostly clean use ["none"]>,
  "degradation_confidence": <0.0-1.0>
}
\end{lstlisting}
Be conservative: only list degradations clearly visible in the image.
\end{promptbox}

\section{Broader Impacts}
\label{sec:supp_impact}
\paragraph{Potential benefits.} Generative Real-ISR is increasingly used to enhance personal photographs, archival material, and images captured under poor conditions. By rewarding outputs that remain supported by the LR evidence, LucidNFT aims to reduce hallucinated text, faces, and structures that viewers may otherwise mistake for recovered detail, and LucidConsistency provides a reference-free signal for flagging outputs whose content departs from the input.

\paragraph{Potential risks.} Improved faithfulness does not make generative SR outputs reliable evidence: the restored details remain samples from a learned prior, and both rewards can be imperfect (Sec.~\ref{sec:con}). Outputs of LucidNFT should therefore not be used for forensic analysis, identification of individuals, or medical decisions without independent verification. More realistic restoration may also make manipulated or fabricated imagery harder to distinguish from genuine photographs. LucidLR contains photographs of people collected from Wikimedia Commons; we screen it for explicit content, personal data, and hate symbols, release it only under licenses that permit redistribution, and record source and license information for every image (Appendix~\ref{sec:supp_lucidlr_release}), but users remain responsible for complying with these licenses and with applicable personality rights.


\clearpage
\section*{NeurIPS Paper Checklist}

\begin{enumerate}

\item {\bf Claims}
    \item[] Question: Do the main claims made in the abstract and introduction accurately reflect the paper's contributions and scope?
    \item[] Answer: \answerYes{}.
    \item[] Justification: The abstract and introduction state the main contributions, including LucidConsistency, decoupled reward normalization, and LucidLR. The experimental sections evaluate these claims through human-aligned faithfulness evaluation, rollout advantage analysis, quantitative comparisons, and ablations.

\item {\bf Limitations}
    \item[] Question: Does the paper discuss the limitations of the work performed by the authors?
    \item[] Answer: \answerYes{}.
    \item[] Justification: The conclusion contains a dedicated Limitations paragraph covering extreme degradations, reward-hacking risk, the domain coverage of LucidLR, the dependency of LucidConsistency on its Qwen3-VL-Embedding backbone, its dual role as reward and evaluation metric, and the absence of an external LR-faithfulness benchmark.
    \item[] Guidelines:
    \begin{itemize}
        \item The answer \answerNA{} means that the paper has no limitation while the answer \answerNo{} means that the paper has limitations, but those are not discussed in the paper.
        \item The authors are encouraged to create a separate ``Limitations'' section in their paper.
        \item The paper should point out any strong assumptions and how robust the results are to violations of these assumptions.
        \item The authors should reflect on the scope of the claims made, e.g., if the approach was only tested on a few datasets or with a few runs.
        \item The authors should reflect on the factors that influence the performance of the approach.
        \item The authors should discuss the computational efficiency of the proposed algorithms and how they scale with dataset size.
        \item If applicable, the authors should discuss possible limitations of their approach to address problems of privacy and fairness.
        \item The authors should use their best judgment and recognize that transparency about limitations helps preserve the integrity of the community.
    \end{itemize}

\item {\bf Theory assumptions and proofs}
    \item[] Question: For each theoretical result, does the paper provide the full set of assumptions and a complete (and correct) proof?
    \item[] Answer: \answerNA{}.
    \item[] Justification: The paper is primarily empirical and does not present formal theoretical results or proofs. The mathematical formulations describe the training objectives and reward normalization procedure.
    \item[] Guidelines:
    \begin{itemize}
        \item The answer \answerNA{} means that the paper does not include theoretical results.
        \item All the theorems, formulas, and proofs in the paper should be numbered and cross-referenced.
        \item All assumptions should be clearly stated or referenced in the statement of any theorems.
        \item The proofs can either appear in the main paper or the supplemental material.
        \item Any informal proof provided in the core of the paper should be complemented by formal proofs provided in appendix or supplemental material.
        \item Theorems and Lemmas that the proof relies upon should be properly referenced.
    \end{itemize}

\item {\bf Experimental result reproducibility}
    \item[] Question: Does the paper fully disclose all the information needed to reproduce the main experimental results of the paper to the extent that it affects the main claims and/or conclusions of the paper?
    \item[] Answer: \answerYes{}.
    \item[] Justification: The paper describes the model backbones, datasets, training settings, reward definitions, rollout settings, evaluation protocols, and baselines used in the experiments. Additional degradation ranges, dataset statistics, and implementation details are provided in the appendix or supplementary material.
    \item[] Guidelines:
    \begin{itemize}
        \item The answer \answerNA{} means that the paper does not include experiments.
        \item If the paper includes experiments, a \answerNo{} answer to this question will not be perceived well by the reviewers.
        \item If the contribution is a dataset and/or model, the authors should describe the steps taken to make their results reproducible or verifiable.
        \item Reproducibility can be provided via code, detailed instructions, hosted models, released checkpoints, or other appropriate means.
        \item If the contribution is primarily a new algorithm, the paper should make it clear how to reproduce that algorithm.
        \item If the contribution is primarily a new model architecture, the paper should describe the architecture clearly and fully.
        \item If the contribution is a new model, there should be a way to access or reproduce this model.
        \item Authors should describe the particular avenue they provide for reproducibility when full reproduction is difficult.
    \end{itemize}

\item {\bf Open access to data and code}
    \item[] Question: Does the paper provide open access to the data and code, with sufficient instructions to faithfully reproduce the main experimental results, as described in supplemental material?
    \item[] Answer: \answerYes{}.
    \item[] Justification: LucidLR is released as a pre-fetched image bundle and as a metadata index with a reference downloader, each image with a JSON sidecar recording its source and license (Appendix~\ref{sec:supp_lucidlr_release}); implementation resources are released together with the dataset.
    \item[] Guidelines:
    \begin{itemize}
        \item The answer \answerNA{} means that paper does not include experiments requiring code.
        \item Please see the NeurIPS code and data submission guidelines for more details.
        \item While release of code and data is encouraged, \answerNo{} is an acceptable answer when justified.
        \item The instructions should contain the exact command and environment needed to reproduce the results.
        \item The authors should provide instructions on data access and preparation.
        \item The authors should provide scripts to reproduce all experimental results for the proposed method and baselines, or state which ones are omitted and why.
        \item At submission time, authors should release anonymized versions if applicable.
        \item Providing as much information as possible in supplemental material is recommended.
    \end{itemize}

\item {\bf Experimental setting/details}
    \item[] Question: Does the paper specify all the training and test details necessary to understand the results?
    \item[] Answer: \answerYes{}.
    \item[] Justification: The implementation details specify the backbones, LoRA settings, training resolutions, learning rates, batch sizes, rollout counts, inference steps, rewards, optimizer, and evaluation datasets. Additional controlled degradation settings are provided in the appendix.
    \item[] Guidelines:
    \begin{itemize}
        \item The answer \answerNA{} means that the paper does not include experiments.
        \item The experimental setting should be presented in the core of the paper to a level of detail necessary to appreciate the results.
        \item Full details can be provided either with the code, in appendix, or as supplemental material.
    \end{itemize}

\item {\bf Experiment statistical significance}
    \item[] Question: Does the paper report error bars suitably and correctly defined or other appropriate information about the statistical significance of the experiments?
    \item[] Answer: \answerNo{}.
    \item[] Justification: The paper reports extensive comparisons across multiple datasets, metrics, and ablations, but does not report error bars or formal statistical significance tests due to the high computational cost of repeated RL fine-tuning runs. Human-aligned evaluation is reported using aggregate agreement metrics.
    \item[] Guidelines:
    \begin{itemize}
        \item The answer \answerNA{} means that the paper does not include experiments.
        \item The authors should answer \answerYes{} if results are accompanied by error bars, confidence intervals, or statistical significance tests.
        \item The factors of variability captured by the error bars should be clearly stated.
        \item The method for calculating error bars should be explained.
        \item The assumptions made should be given.
        \item It should be clear whether the error bar is the standard deviation or the standard error of the mean.
        \item It is OK to report 1-sigma error bars, but one should state it.
        \item For asymmetric distributions, authors should avoid misleading symmetric error bars.
        \item If error bars are reported, the text should explain how they were calculated and reference the corresponding figures or tables.
    \end{itemize}

\item {\bf Experiments compute resources}
    \item[] Question: For each experiment, does the paper provide sufficient information on the computer resources needed to reproduce the experiments?
    \item[] Answer: \answerYes{}.
    \item[] Justification: The paper reports the GPU type and number used for LucidConsistency and LucidNFT training, together with training steps, batch sizes, rollout counts, and training resolutions. These details provide the main compute requirements for reproducing the reported experiments.
    \item[] Guidelines:
    \begin{itemize}
        \item The answer \answerNA{} means that the paper does not include experiments.
        \item The paper should indicate the type of compute workers, CPU or GPU, internal cluster or cloud provider, including relevant memory and storage.
        \item The paper should provide the compute required for individual experimental runs and estimate the total compute.
        \item The paper should disclose whether the full research project required more compute than the experiments reported in the paper.
    \end{itemize}

\item {\bf Code of ethics}
    \item[] Question: Does the research conducted in the paper conform, in every respect, with the NeurIPS Code of Ethics \url{https://neurips.cc/public/EthicsGuidelines}?
    \item[] Answer: \answerYes{}.
    \item[] Justification: The work uses public or standard research datasets and includes filtering procedures for the newly collected real-world degraded images. The human annotation protocol is limited to image quality and faithfulness judgments and does not collect sensitive personal information.
    \item[] Guidelines:
    \begin{itemize}
        \item The answer \answerNA{} means that the authors have not reviewed the NeurIPS Code of Ethics.
        \item If the authors answer \answerNo{}, they should explain the special circumstances that require a deviation from the Code of Ethics.
        \item The authors should make sure to preserve anonymity where required.
    \end{itemize}

\item {\bf Broader impacts}
    \item[] Question: Does the paper discuss both potential positive societal impacts and negative societal impacts of the work performed?
    \item[] Answer: \answerYes{}.
    \item[] Justification: Appendix~\ref{sec:supp_impact} discusses the potential benefit of more LR-faithful generative restoration and the risks that restored details are mistaken for reliable evidence, that more realistic restoration hampers the detection of manipulated imagery, and that LucidLR contains photographs of people, together with the screening and licensing measures we apply.
    \item[] Guidelines:
    \begin{itemize}
        \item The answer \answerNA{} means that there is no societal impact of the work performed.
        \item If the authors answer \answerNA{} or \answerNo{}, they should explain why.
        \item Examples of negative societal impacts include malicious or unintended uses, fairness considerations, privacy considerations, and security considerations.
        \item Foundational research not tied to deployment may still have direct paths to negative applications.
        \item Authors should consider harms from intended use, incorrect results, and misuse.
        \item If there are negative societal impacts, authors may discuss mitigation strategies.
    \end{itemize}

\item {\bf Safeguards}
    \item[] Question: Does the paper describe safeguards that have been put in place for responsible release of data or models that have a high risk for misuse?
    \item[] Answer: \answerYes{}.
    \item[] Justification: For LucidLR, every image is screened in two passes by a vision-language model (DeepSeek-V4.1-Flash), and images with explicit sexual content, photographic gore, personal data, or hate symbols are removed (Appendix~\ref{sec:supp_lucidlr_release}). Only images whose Commons licenses permit redistribution are included, and images with uncertain redistribution status are kept as URL-only entries.
    \item[] Guidelines:
    \begin{itemize}
        \item The answer \answerNA{} means that the paper poses no such risks.
        \item Released models with high misuse or dual-use risk should be released with safeguards such as usage restrictions or safety filters.
        \item Scraped datasets could pose safety risks, and authors should describe how unsafe images were avoided.
        \item Providing effective safeguards is challenging, but authors are encouraged to make a best-faith effort.
    \end{itemize}

\item {\bf Licenses for existing assets}
    \item[] Question: Are the creators or original owners of assets used in the paper properly credited and are the license and terms of use explicitly mentioned and properly respected?
    \item[] Answer: \answerYes{}.
    \item[] Justification: The paper cites the datasets, models, and methods used, including RealSR, DRealSR, RealLQ250, LSDIR, Real-ESRGAN, Qwen3-VL-Embedding, DiffusionNFT, and the evaluated SR baselines. LucidLR is collected from Wikimedia Commons; each image keeps its original license (mainly CC~BY and CC~BY-SA, plus public-domain and CC0 files), and author and license information is recorded per image for attribution.
    \item[] Guidelines:
    \begin{itemize}
        \item The answer \answerNA{} means that the paper does not use existing assets.
        \item The authors should cite the original paper that produced the code package or dataset.
        \item The authors should state which version of the asset is used and, if possible, include a URL.
        \item The name of the license should be included for each asset.
        \item For scraped data from a particular source, the copyright and terms of service should be provided.
        \item If assets are released, license, copyright information, and terms of use should be provided.
        \item For repackaged datasets, both the original and derived licenses should be provided.
        \item If licensing information is unavailable online, authors are encouraged to contact asset creators.
    \end{itemize}

\item {\bf New assets}
    \item[] Question: Are new assets introduced in the paper well documented and is the documentation provided alongside the assets?
    \item[] Answer: \answerYes{}.
    \item[] Justification: The paper introduces LucidLR and describes its source, scale, screening pipeline, licensing, release format, per-image metadata schema, and intended usage, together with semantic, resolution, and degradation statistics (Appendices~\ref{sec:supp_lucidlr_release} and~\ref{sec:supp_lucidlr_stats}).
    \item[] Guidelines:
    \begin{itemize}
        \item The answer \answerNA{} means that the paper does not release new assets.
        \item Researchers should communicate dataset, code, and model details via structured templates.
        \item The paper should discuss whether and how consent was obtained from people whose asset is used.
        \item At submission time, assets should be anonymized if applicable.
    \end{itemize}

\item {\bf Crowdsourcing and research with human subjects}
    \item[] Question: For crowdsourcing experiments and research with human subjects, does the paper include the full text of instructions given to participants and screenshots, if applicable, as well as details about compensation?
    \item[] Answer: \answerYes{}.
    \item[] Justification: The paper describes both human studies, the evaluator-alignment ranking study and the model-level pairwise preference study, including the LR-faithfulness instructions, the candidate presentation format, the number of annotators and judgments, and, for the model-level study, inter-annotator agreement (Fleiss' $\kappa$). Full annotation instructions and interface details are provided in the appendix.
    \item[] Guidelines:
    \begin{itemize}
        \item The answer \answerNA{} means that the paper does not involve crowdsourcing nor research with human subjects.
        \item Including this information in the supplemental material is fine.
        \item According to the NeurIPS Code of Ethics, workers involved in data collection, curation, or other labor should be paid at least the minimum wage in the country of the data collector.
    \end{itemize}

\item {\bf Institutional review board (IRB) approvals or equivalent for research with human subjects}
    \item[] Question: Does the paper describe potential risks incurred by study participants, whether such risks were disclosed to the subjects, and whether IRB approvals or equivalent approval/review were obtained?
    \item[] Answer: \answerNA{}.
    \item[] Justification: The human study consists of expert judgments on image restoration faithfulness and does not collect personal, sensitive, or behavioral data from participants. The study follows the authors' institutional requirements for low-risk annotation studies.
    \item[] Guidelines:
    \begin{itemize}
        \item The answer \answerNA{} means that the paper does not involve crowdsourcing nor research with human subjects.
        \item Depending on the country in which research is conducted, IRB approval or equivalent may be required for any human subjects research.
        \item Authors are expected to adhere to the NeurIPS Code of Ethics and their institutional guidelines.
        \item For initial submissions, do not include information that would break anonymity.
    \end{itemize}

\item {\bf Declaration of LLM usage}
    \item[] Question: Does the paper describe the usage of LLMs if it is an important, original, or non-standard component of the core methods in this research?
    \item[] Answer: \answerNA{}.
    \item[] Justification: LLMs are not an important, original, or non-standard component of the proposed method. A vision-language model (DeepSeek-V4.1-Flash) is used only for dataset curation and characterization, namely content screening and semantic/degradation annotation of LucidLR, and this use is described in Sec.~\ref{sec:lucidlr} and Appendices~\ref{sec:supp_lucidlr_release}--\ref{sec:supp_lucidlr_stats}.
    \item[] Guidelines:
    \begin{itemize}
        \item The answer \answerNA{} means that the core method development in this research does not involve LLMs as any important, original, or non-standard components.
        \item Please refer to the NeurIPS LLM policy in the handbook for what should or should not be described.
    \end{itemize}

\end{enumerate}

\end{document}